\documentclass[11pt,letterpaper]{article}
\usepackage[in]{fullpage}

\usepackage[authoryear,round]{natbib}
\usepackage{microtype}
\usepackage{graphicx}
\usepackage{booktabs} 
\usepackage{amsmath,amssymb} 
\usepackage{subcaption}
\usepackage[font=small]{caption}

\usepackage{algorithm}
\usepackage{algorithmic}

\title{On the Effectiveness of Adversarial Training\\Against Common Corruptions}

\usepackage{microtype}
\usepackage{graphicx}
\usepackage{booktabs} 
\usepackage{amsmath,amssymb} 
\usepackage{subcaption}

\usepackage[colorlinks=true,citecolor=blue,urlcolor=red]{hyperref}  
\usepackage{nicefrac}
\usepackage[table]{xcolor}
\usepackage{arydshln}
\usepackage{multirow}
\usepackage{wrapfig}

\newcommand{\comment}[1]{}

\usepackage[linesnumbered,algoruled,boxed,lined,noend,algo2e]{algorithm2e}  

\def\minop{\mathop{\rm min}\limits}
\def\maxop{\mathop{\rm max}\limits}

\def\R{\mathbb{R}}

\newcommand{\myparagraph}{\textbf}

\newcommand{\norm}[1]{\left\|#1\right\|}

\newcommand*{\defeq}{\stackrel{\text{def}}{=}}

\DeclareMathOperator{\eps}{\mathbf{\varepsilon}}
\DeclareMathOperator{\lpips}{d_{\text{LPIPS}}}

\DeclareMathOperator{\E}{\mathbb{E}}
\let\l\relax
\DeclareMathOperator{\l}{\mathbf{\ell}}

\makeatletter
\def\blfootnote{\xdef\@thefnmark{}\@footnotetext}
\makeatother

\usepackage{enumitem}

\author{
	Klim Kireev\textsuperscript{*}\\
	EPFL\\
	\and
	Maksym Andriushchenko\textsuperscript{*}\\
	EPFL\\
	\and
	Nicolas Flammarion \\
	EPFL\\
}

\date{\vspace{-5ex}}

\begin{document}
	
	\blfootnote{\vspace{1em} \textsuperscript{*}Equal contribution.}
	\maketitle

	\begin{abstract}
		The literature on robustness towards common corruptions shows no consensus on whether adversarial training can improve the performance in this setting. First, we show that, when used with an appropriately selected perturbation radius, $\l_p$ adversarial training can serve as a strong baseline against common corruptions improving both accuracy and calibration. Then we explain why adversarial training performs better than data augmentation with simple Gaussian noise which has been observed to be a meaningful baseline on common corruptions. Related to this, we identify the \textit{$\sigma$-overfitting} phenomenon when Gaussian augmentation overfits to a particular standard deviation used for training which has a significant detrimental effect on common corruption accuracy. We discuss how to alleviate this problem and then how to further enhance $\l_p$ adversarial training by introducing an \textit{efficient relaxation} of adversarial training with \textit{learned perceptual image patch similarity} as the distance metric. Through experiments on CIFAR-10 and ImageNet-100, we show that our approach does not only improve the $\l_p$ adversarial training baseline but also has cumulative gains with data augmentation methods such as AugMix, DeepAugment, ANT, and SIN, leading to state-of-the-art performance on common corruptions.
		The code of our experiments is publicly available at \url{https://github.com/tml-epfl/adv-training-corruptions}.
	\end{abstract}

	\section{Introduction}
	
	\begin{wrapfigure}{r}{0.445\textwidth}
		\vspace{-10mm}
		\begin{center}
			\includegraphics[width=0.44\columnwidth]{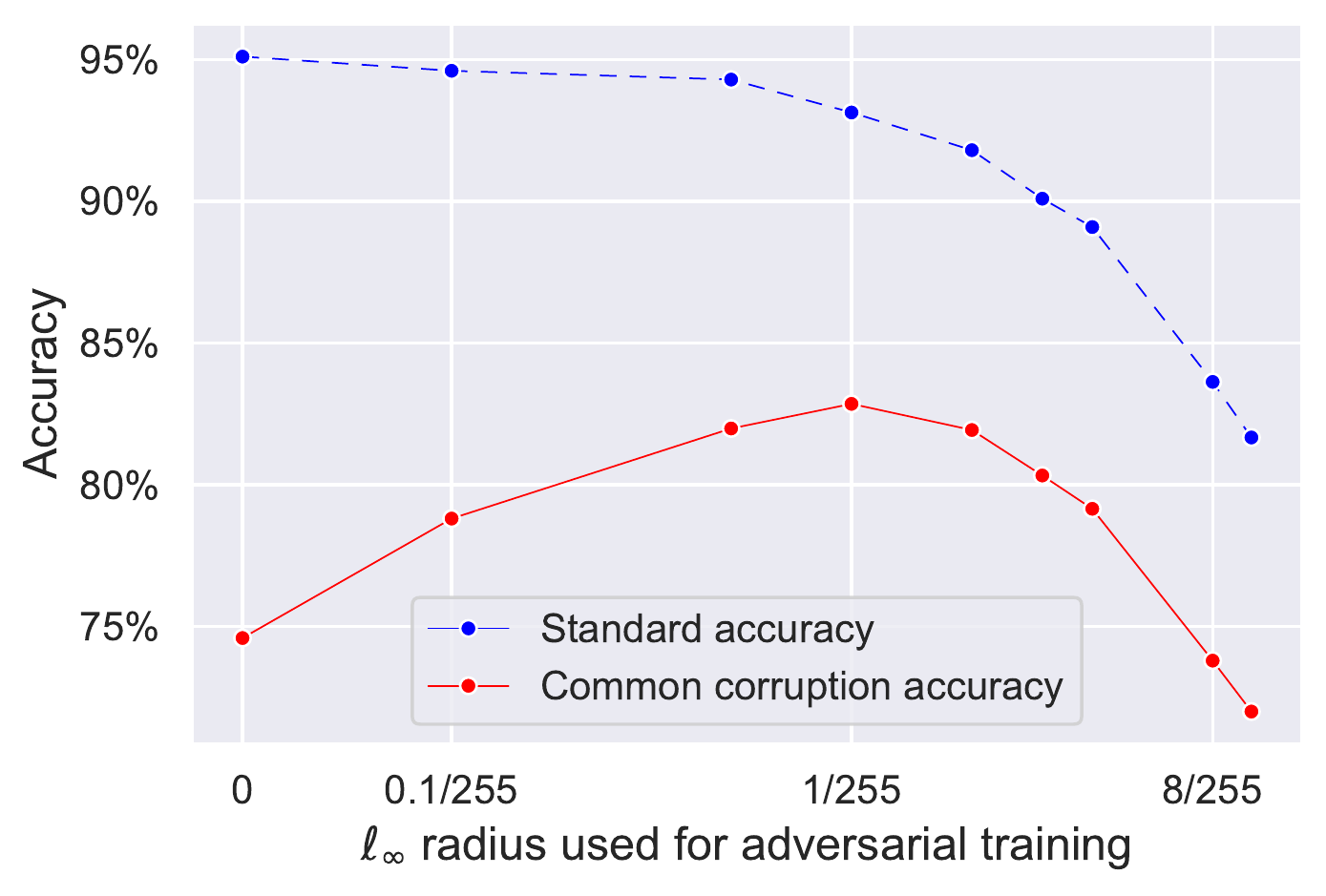}
			\vskip -0.1in
			\caption{
				Accuracy on common corruptions from CIFAR-10-C for ResNet-18 models adversarially trained using different $\l_\infty$ radii. We observe that the performance with $\eps=\nicefrac{1}{255}$ is significantly higher than with the standardly used $\eps=\nicefrac{8}{255}$.
			} 
			\label{fig:linf_eps_c10c_acc}
		\end{center}
		\vspace{-12mm}
	\end{wrapfigure}
	Despite achieving human-level performance on many computer vision tasks, deep neural networks are still not as robust as humans towards various distribution shifts \citep{szegedy2013intriguing, taori2020measuring} including common image corruptions \citep{hendrycks2019benchmarking}. 
	Attempts to understand the vulnerability towards such shifts include analysis of the network architecture \citep{azulay2019deep},
	the features contained in the data \citep{ilyas2019adversarial}, and frequency analysis of neural networks \citep{yin2019fourier, ortiz2020hold}.
	Many approaches have been suggested to improve their robustness to these shifts including approaches based on data augmentations \citep{cubuk2018autoaugment, hendrycks2019augmix}, adversarial training \citep{madry2018towards, laidlaw2021perceptual}, and pretraining \citep{hendrycks2019using}.
	
	Although data augmentation methods tend to improve the performance under common synthetic corruptions \citep{hendrycks2019augmix}, these augmentations are often ad hoc and may have substantial overlap with the corruptions evaluated at test time. 
	At the same time, there is a large amount of literature on adversarial training with $\l_p$-bounded perturbations \citep{goodfellow2014explaining,madry2018towards}.
	Adversarial training emerged as a principled approach to improve the worst-case performance of the model against \textit{small} $\l_p$ perturbations.
	However, common image corruptions have a very high $\l_p$ distance from clean samples, so the utility of using $\l_p$ adversarial training for them is not obvious.
	This leads us to explore the following question:
	\begin{center}
		\textit{How can we improve the performance on common image corruptions using adversarial training?}
	\end{center}
	
	We make the following contributions in our paper:
	\begin{itemize}
		\vspace{-.1cm}
		\item We show that $\l_p$ adversarial training with an \textit{appropriately selected} perturbation radius can serve as a strong baseline against common image corruptions improving both accuracy and calibration on corrupted images.
		\vspace{-.1cm}
		\item We analyze the success of $\l_p$ adversarial training via a comparison to other natural baselines such as Gaussian data augmentation. 
		We observe that it can overfit to the perturbation size it has been trained which, however, does not happen for adversarial training.
		\vspace{-.1cm}
		\item We introduce an efficient relaxation of adversarial training with \textit{learned perceptual image patch similarity} (LPIPS) \citep{zhang2018unreasonable} based on layerwise adversarial perturbations. This new relaxation is at least as effective as previous approaches \citep{laidlaw2021perceptual} but significantly faster to train. 
		\vspace{-.1cm}
		\item We show that our relaxation approach has cumulative gains with existing data augmentation methods such as AugMix, DeepAugment, ANT, and SIN leading to state-of-the-art performance on common corruptions from CIFAR-10-C and ImageNet-100-C.
	\end{itemize}

	\section{Related work}
	We provide here an overview of relevant works on common image corruptions, different data augmentation methods proposed to improve the performance on corruptions, and then we discuss papers on adversarial robustness with respect to both $\l_p$ and non-$\l_p$ perturbations.
	
	\myparagraph{Common image corruptions.}
	\cite{dodge2017study} first find that despite being on par with the human vision on standard images, deep networks perform suboptimally on common corruptions such as noise and blur.
	\cite{geirhos2018generalisation} measure the performance of deep networks on $12$ different image corruption types but find that data augmentation on one type of corruption does not tend to improve the performance on others. 
	However, these findings are reconsidered in \cite{rusak2020simple} where Gaussian data augmentation is shown to help for a wide range of image corruptions.
	In a standardization effort, \cite{hendrycks2019benchmarking} introduce a few image classification datasets---in particular, CIFAR-10-C and ImageNet-C---with 15 different common corruptions from four categories: noise, blur, weather, and digital corruptions.
	\cite{ovadia2019can} show that not only acccuracy but also calibration deteriorates under these common corruptions.
	%
	\citep{schneider2020improving, nandy2021adversarially} show that robustness to common corruptions can be improved by using test-time adaptation, e.g., via recomputing the batch normalization statistics.
	\cite{radford2021learning} show that contrastive pretraining on a very large set of image-caption pairs can substantially improve robustness on various distribution shifts including common corruptions.

	\myparagraph{Data augmentations.}
	Data augmentation is a widely used technique to improve the generalization. 
	Besides classical image transformations like random flipping or cropping, many other approaches have been proposed such as linearly interpolating between images and their labels \citep{zhang2018mixup}, replacing a part of the image with either a black-colored patch \citep{devries2017improved} or a part of another image \citep{yun2019cutmix}.
	%
	%
	One of the best-performing methods in terms of accuracy and calibration on common corruptions is AugMix \citep{hendrycks2019augmix}, which combines carefully selected augmentations with a regularization term based on the Jensen-Shannon divergence. 
	%
	%
	\cite{taori2020measuring} observe that improvements on synthetic distribution shifts (such as common corruptions) do not necessarily transfer to real distribution shifts. 
	However, \cite{hendrycks2020many} show an example when improving robustness against synthetic blurs also helps against naturally obtained blurred images.

	\myparagraph{$\l_p$ adversarial robustness.}
	Adversarial training in deep learning has been first considered in \cite{goodfellow2014explaining} and later framed as a robust optimization problem by \cite{madry2018towards}.
	%
	%
	The view that adversarial training damages or at least does not improve the performance on \textit{common corruptions} has been prevalent in the literature \citep{hendrycks2019augmix, rusak2020simple, hendrycks2020many}.
	However, previous works directly use publicly available robust models without adjusting the perturbation radius used for adversarial training. 
	For example, \citet{rusak2020simple} show that adversarially trained ImageNet models from \cite{xie2019feature}, \cite{shafahi2019adversarial}, and \cite{shafahi2020universal} do not help on ImageNet-C compared to standardly trained models.
	However, \cite{ford2019adversarial} report that $\l_\infty$ adversarially trained models on CIFAR-10 from \cite{madry2018towards} do lead to an improvement on CIFAR-10-C compared to a standard model.
	The approach of \cite{xie2020adversarial}, AdvProp, relies on $\l_\infty$ adversarial training to improve standard and corruption accuracy but they advocate the use of \textit{auxiliary} batch normalization layers for standard and adversarial training examples. We find that similar performance can be achieved on common corruptions using vanilla adversarial training without a customized use of BatchNorm layers.
	\cite{kang2019transfer} study the robustness transfer between $\l_p$-robust models and \textit{adversarially optimized} elastic and JPEG corruptions. They show that $\l_p$ adversarial training can increase robustness against these two types of adversarial perturbations, but robustness does not transfer in all the cases and sometimes may even hurt robustness against other perturbation types. 
	%
	%

	\myparagraph{Non-$\l_p$ adversarial robustness.}
	\cite{volpi2018generalizing} propose Lagrangian-style adversarial training in the input space and in the last layer of the network. 
	\cite{stutz2019disentangling} propose \textit{on-manifold} adversarial training which is performed in the latent space of a VAE-GAN generative model. However, its success crucially depends on the quality of the generative model which could not be scaled beyond simple image recognition datasets.
	%
	%
	%
	\cite{wei2020improved} derive generalization bounds that motivate adversarial training with respect to all network layers which they use to improve $\l_p$ robustness. 
	Recently, \cite{laidlaw2021perceptual} provided algorithms for approximate \textit{perceptual adversarial training} based on the LPIPS distance \citep{zhang2018unreasonable} which is defined via activations of a neural network. They aim at improving robustness against new types of adversarial perturbations that were unseen during training.

	\section{$\ell_p$ adversarial training improves the performance on common corruptions}
	Here we formally introduce adversarial training and show that it can lead to non-trivial improvements in accuracy and calibration on common corruptions.
	
	\myparagraph{Background on adversarial training.}
	Let $\l(x,y;\theta)$ denote the loss of a classifier parametrized by $\theta \in \R^m $ on the sample $(x,y)\sim D$ where $D$ is the data distribution.
	Previous works \citep{shaham2015understanding,madry2018towards} formalized the goal of training adversarially robust models as the following optimization problem: 
	\begin{align}
	\minop_{\theta} \E_{(x, y) \sim D} \big[ \maxop_{\delta\in\Delta} \l(x+\delta, y; \theta) \big].
	\label{eq:rob_opt_general}
	\end{align}
	In this section, we focus on the $\ell_p$ threat model, i.e. $\Delta = \{\delta\in\R^d: \norm{\delta}_p \leq \varepsilon, \ x + \delta \in [0, 1]^d\}$, where the adversary can change each input $x$ in an $\eps$-ball around it while making sure that the input $x+\delta$ does not exceed its natural range.
	A common way to solve the inner maximization problem is the \textit{projected gradient descent} method (PGD) defined by the following recursion initialized at $\delta^{(0)}$:
	\begin{align} \label{eq:pgd-def}
	\delta^{(t+1)} \defeq \Pi_{\Delta} \left[\delta^{(t)} + \alpha \nabla_{\delta^{(t)}} \l(x+\delta^{(t)}, y; \theta)\right],
	\end{align}
	where $\Pi$ is the projection operator on the set $\Delta$, and $\alpha$ is the step size of PGD. 
	Instead of the gradient, one often uses the gradient sign update for $\l_\infty$ perturbations or the $\l_2$ normalized update for $\l_2$ perturbations.
	$\delta^{(0)}$ can be initialized as any point inside $\Delta$, e.g. as zero,
	or randomly \citep{madry2018towards}.
	
	The one-iteration variant of PGD  is known as the \textit{fast gradient method} (FGM) when the normalized $\l_2$ update is used and as the \textit{fast gradient sign method} (FGSM) when the $\l_\infty$ sign update is used \citep{goodfellow2014explaining}. 
	Note that in both cases the step size is $\alpha=\eps$ which leads to perturbations located on the boundary of the set $\Delta$.
	These methods are fast but sometimes prone to \textit{catastrophic overfitting} when the model overfits to FGM/FGSM but is not robust to iterative PGD attacks \citep{tramer2018ensemble, wong2020fast}.
	This problem can be alleviated by specific regularization methods like CURE \citep{moosavi2019robustness,huang2020bridging} or GradAlign \citep{andriushchenko2020understanding}. 
	However, for small enough $\eps$, adversarial training with FGM/FGSM works as well as multi-step PGD \citep{andriushchenko2020understanding}.

	\myparagraph{Experimental details.}
	We do experiments on two common image classification datasets: CIFAR-10 \citep{krizhevsky2009learning} which has $32\times32$ images, and ImageNet-100 \citep{russakovsky2015imagenet} with $224\times224$ images where we take each tenth class following \cite{laidlaw2021perceptual}. We choose ImageNet-100 since we always perform a grid search over the main hyperparameters such as the perturbation radius for adversarial training 
	which would be too expensive to do on the full ImageNet. Unless mentioned otherwise, we use PreAct ResNet-18 architecture \citep{he2016identity}. We specify the exact hyperparameters in App.~\ref{sec:app_exp_details}. We evaluate the accuracy on common corruptions using CIFAR-10-C and ImageNet-C datasets from \citep{hendrycks2019benchmarking} which contain 15 different synthetic corruptions in 4 categories: blur, noise, digital, weather corruptions. We report the accuracy by averaging over all 5 severity levels. 
	The code of our experiments is publicly available\footnote{\url{https://github.com/tml-epfl/adv-training-corruptions}}.

	\myparagraph{Adversarial training improves accuracy and calibration.} 
	\begin{table}[t]
		\centering
		\small
		\begin{tabular}{@{}lccc}
			& \textbf{Standard} & \textbf{Corruption} & \textbf{Corruption} \\
			\textbf{Training} & \textbf{accuracy} & \textbf{accuracy}   & \textbf{calibration error}  \\
			\midrule
			& \multicolumn{3}{c}{\textbf{CIFAR-10}} \\
			\cmidrule(lr){2-4}
			Standard & 95.1\% & 74.6\% & 16.6\% \\
			$\l_\infty$ adversarial  & 93.3\% & 82.7\% & 10.8\% \\
			$\l_2$ adversarial & 93.6\% & \textbf{83.4\%} & \textbf{10.5\%} \\
			\midrule
			& \multicolumn{3}{c}{\textbf{ImageNet-100}} \\
			\cmidrule(lr){2-4}
			Standard & 86.6\% & 47.5\% & 10.0\% \\
			$\l_\infty$ adversarial  & 86.5\% & 47.7\% & 12.4\% \\
			$\l_2$ adversarial & 86.3\% & \textbf{48.4\%} & \textbf{9.4\%}\\
		\end{tabular}
		\caption{Accuracy and calibration of ResNet-18 models trained on CIFAR-10 and ImageNet-100. $\l_\infty$ and $\l_2$ adversarial training substantially improves accuracy and calibration error (ECE) on corrupted samples.
		}
		\label{tab:at_helps}
	\end{table}
	%
	%
	We start by showing in Fig.~\ref{fig:linf_eps_c10c_acc} the common corruption accuracy of $\l_\infty$ adversarially trained models as it is the most widely studied setting \citep{madry2018towards} and has been reported multiple times in common corruption literature \citep{hendrycks2019augmix, ford2019adversarial, rusak2020simple}.
	Since we are interested primarily in small-$\eps$ adversarial training, we rely throughout the paper on FGM/FGSM for $\l_2$/$\l_\infty$ norms respectively to solve the inner maximization problem~\eqref{eq:rob_opt_general} which only leads to a $2\times$ computational overhead. 
	Note however that we exceptionally use PGD with $10$ steps for $\eps \in \{\nicefrac{8}{255}, \nicefrac{10}{255}\}$ to prevent catastrophic overfitting 
	and allow a direct comparison with previous works.
	We observe that \textit{for the small-$\eps$ regime} around $\eps=\nicefrac{1}{255}$, we get a significant improvement in corruption accuracy: 74.5\% accuracy is achieved with standard training, 82.7\% with adversarial training using $\eps=\nicefrac{1}{255}$, and 73.8\% using the standardly reported threshold $\eps_\infty=\nicefrac{8}{255}$.\footnote{The exact numbers differ from \citep{ford2019adversarial} since we use ResNet-18 instead of WRN-28-10 and different hyperparameters.}
	The reason is that the tradeoff between robustness and accuracy \citep{tsipras2018robustness} has to be carefully balanced---if the standard accuracy drops for higher $\eps$, the corruption accuracy also deteriorates. Thus, selecting the most robust $\l_p$-model does not lead to the optimal performance on common corruptions.
	Alternatively, one can also balance this tradeoff by mixing clean and adversarial samples, but it overall leads to similar results (see App.~\ref{sec:app_ablation_advprop_25_50_75_100} for details), so we focus on adversarial training with 100\% adversarial samples for the rest of the paper.
	
	\begin{wrapfigure}{r}{0.43\textwidth}
		\vspace{-9.5mm}
		\begin{center}
			\includegraphics[width=0.435\columnwidth]{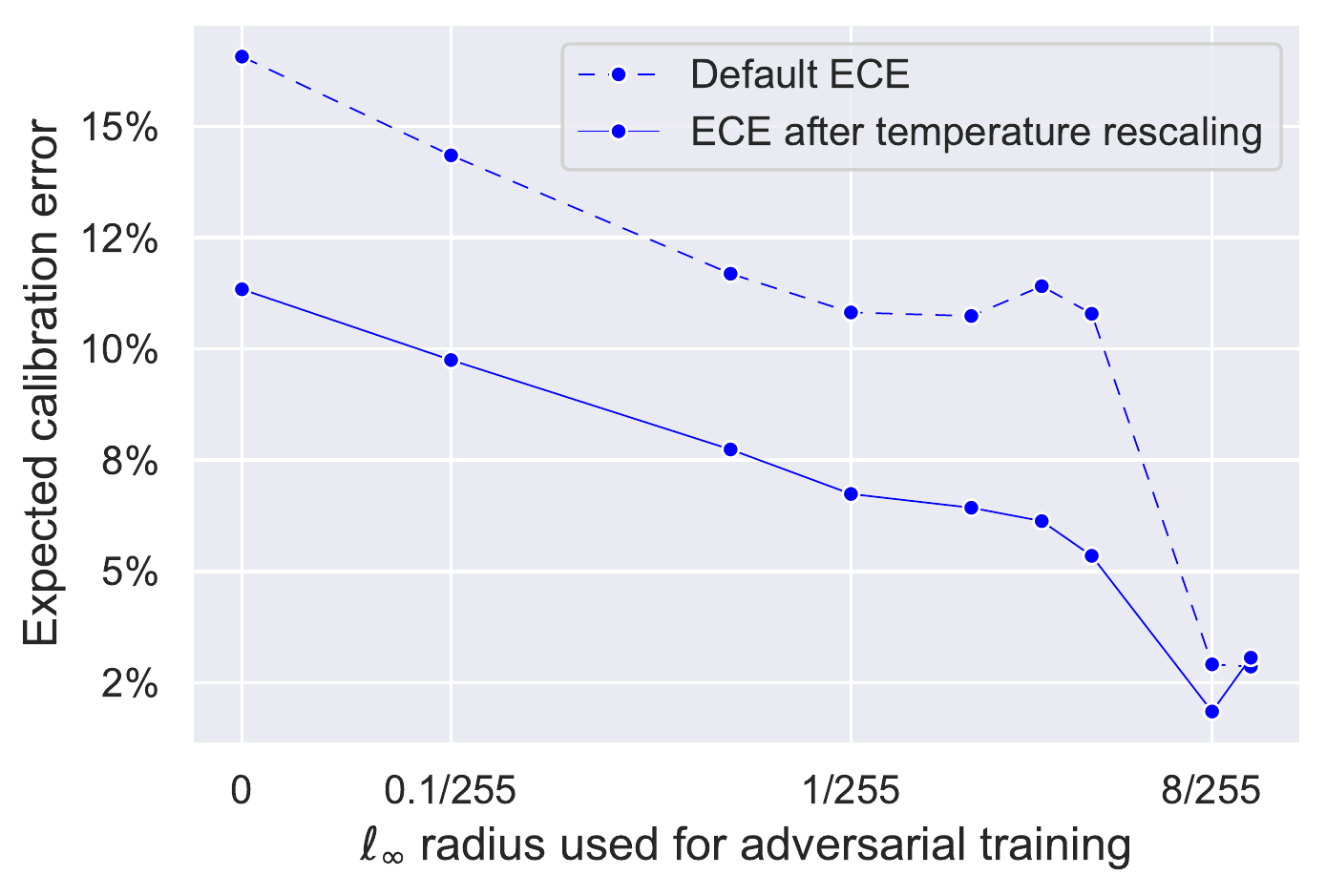}
			\vspace{-6.5mm}
			\caption{Expected calibration error on CIFAR-10-C for $\l_\infty$ adversarially trained models.}
			\label{fig:calibration_linf}
		\end{center}
		\vspace{-7mm}
	\end{wrapfigure}
	Additionally, we show that predicted probabilities of adversarially trained models are significantly better \textit{calibrated on common corruptions}. 
	We believe that calibration is another important aspect of the model's trustworthiness, which is particularly important in the presence of out-of-distribution data such as corrupted images.
	In Fig.~\ref{fig:calibration_linf}, we plot the expected calibration error (ECE) \citep{guo2017calibration} on CIFAR-10-C for models trained with different $\l_\infty$-radii.
	We observe that the ECE---both with and without temperature rescaling (see App.~\ref{sec:app_calibration} for details)---follows a decreasing trend over $\l_\infty$-radii which is expected since a classifier that predicts uniform probabilities over classes is perfectly calibrated.
	In particular, the most accurate model trained with $\eps_\infty=\nicefrac{1}{255}$ has a much lower ECE than the standard model: 10.8\% instead of 16.6\%, and with temperature rescaling 6.7\% instead of 11.3\%. 
	
	We further compare the performance in the $\l_2$ perturbation model. In Table~\ref{tab:at_helps}, we report results of standard, $\l_\infty$, and $\l_2$ adversarial training on CIFAR-10 and ImageNet-100 where we perform a detailed grid search for each model over the perturbation radius $\eps$. To the best of our knowledge, we show for the first time that adversarial training improves calibration (see also App.~\ref{sec:app_calibration}) while increasing the accuracy and that it helps on ImageNet-C, and not only on CIFAR-10-C. We generally observe that $\l_2$ adversarial training performs better than $\l_\infty$, thus we focus on it in the next section.

	\section{Understanding the effect of adversarial training on image corruptions}
	Here we compare $\l_2$ adversarial training to other natural baselines 
	and discuss the main conceptual differences.
	
	\label{sec:at_vs_gauss}
	\begin{wrapfigure}{r}{0.43\textwidth}
		\vspace{-10mm}
		\begin{center}
			\includegraphics[width=0.43\columnwidth]{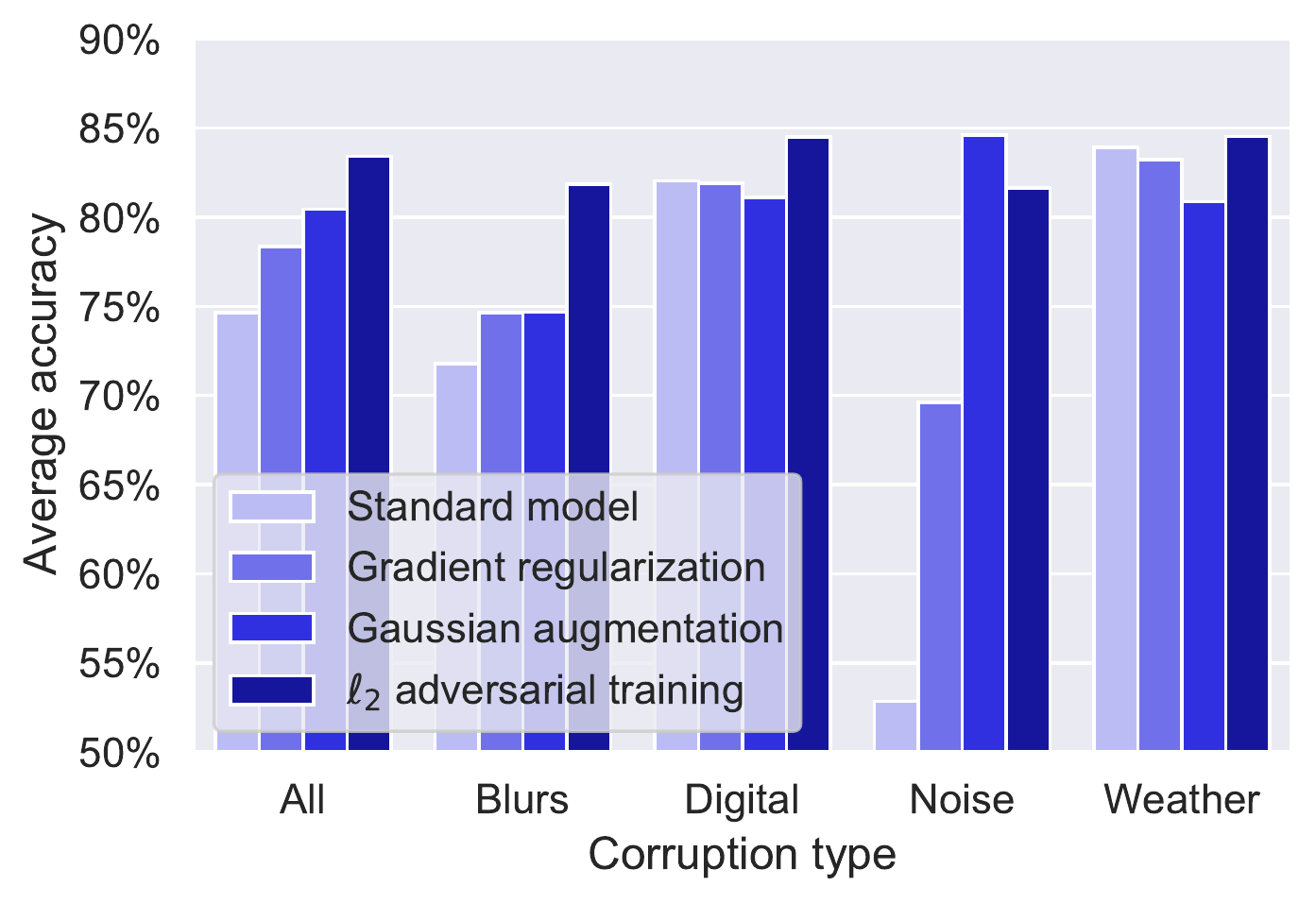}
			\vspace{-7mm}
			\caption{Accuracy for different corruption types on CIFAR-10-C. Unlike other methods, adversarial training improves the performance on each corruption type compared to standard training.}
			\label{fig:barplot_corruption_types}
		\end{center}
		\vspace{-8mm}
	\end{wrapfigure}
	
	\myparagraph{Comparing natural baselines across corruption types.}
	We compare $\ell_2$ adversarial training with a few simple baselines: standard training, gradient regularization \citep{drucker1992improving}, and standard Gaussian data augmentation. To ensure a fair comparison, we perform a grid search for each method over the perturbation radius $\eps$, regularization parameter $\lambda$, and noise standard deviation $\sigma$ respectively.
	We choose to compare to gradient regularization since it is an established regularization method that may have a similar effect to adversarial training with small perturbations \citep{simon2019first}.
	We aggregate the corruptions over each type (blurs, digital, noise, weather) and plot the results in Fig.~\ref{fig:barplot_corruption_types} and report results over each corruption in Fig.~\ref{fig:barplot_corruptions} in the Appendix.
	
	First, we observe that adversarial training is the best performing method
	%
	and that unlike other methods, $\l_2$ adversarial training helps for \textit{each} corruption type. 
	At the same time, Gaussian augmentation \textit{degrades} the performance on digital and weather corruptions while very significantly improving the performance for noise corruptions which is expected as the Gaussian noise used for training is also contained in the noise corruptions. 
	Interestingly, for the fog and contrast corruptions, the performance degrades for \textit{all} methods (see Table~\ref{tab:clean_gradreg_gauss_adv} in App.~\ref{sec:add_exps_app}), consistently with the observation made in \cite{ford2019adversarial}. 
	Our results also suggest that the impact of gradient regularization is limited and it cannot explain the accuracy gains of both adversarial training and Gaussian augmentation as one could expect from the fact that these methods are equivalent to gradient regularization when used with \textit{sufficiently} small parameters $\sigma$ and $\eps$ \citep{bishop1995training}.

	\myparagraph{Worst-case vs average-case behavior.}
	\cite{ford2019adversarial} show that the robustness to Gaussian noise and adversarial perturbations are closely related. More precisely, they show using concentration of measure arguments that a non-zero error rate under Gaussian perturbation implies the existence of small adversarial perturbations and consequently that improving adversarial robustness leads to an improvement in robustness against Gaussian perturbations. This finding is consistent with what we observe here.
	What remains to be understood is why adversarial training performs \textit{better} than Gaussian augmentation on common corruptions. The main difference between both methods appears when analyzing the objectives that both methods minimize. For a single sample $x$, the loss function considered in Gaussian augmentation is:
	\begin{align*}
	\mathbb{E}_{d \sim N(0, I\sigma^2)} \left[ \l(\theta, x + d) \right]  \  \sim  \  \mathbb{E}_{\rho: ||\rho||_2 = \sigma\sqrt{d}} \left[ \l(\theta, x + \rho) \right],
	\end{align*}
	since Gaussian vectors with variance $\sigma^2 I$ are highly concentrated on the sphere of radius $\sigma \sqrt{d}$ in high dimensions. Therefore Gaussian augmentation amounts to minimize an \emph{averaged} objective where perturbations are averaged over the \textit{sphere}. However, the objective behind adversarial training defined in Eq.~\eqref{eq:rob_opt_general} amounts to minimize a \emph{worst-case} loss based on the worst-case perturbation in the \textit{ball}. The key difference is that  minimization of the expected value of the loss function \emph{does not guarantee} any behavior inside the sphere.
	
\begin{figure}[t]
		\centering
		\begin{minipage}{.48\textwidth}
			\centering
			\includegraphics[width=0.9\columnwidth]{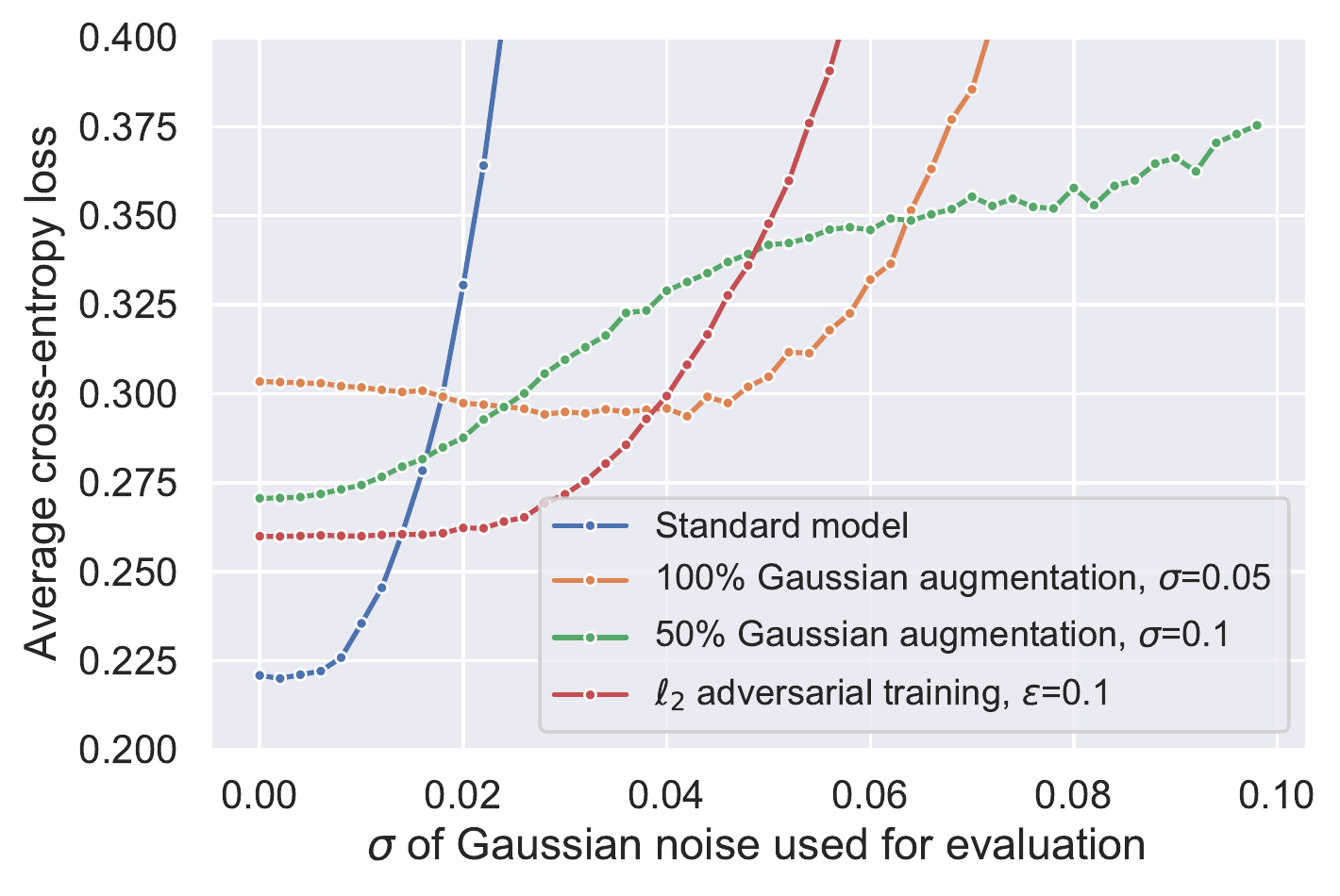}
			\vspace{-1mm}
			\caption{Average cross-entropy loss under Gaussian noise for different training methods. 
			}
			\label{fig:sigma_overfitting}
		\end{minipage}
		\hspace{2mm}
		\begin{minipage}{.48\textwidth}
			\centering
			\includegraphics[width=0.485\columnwidth]{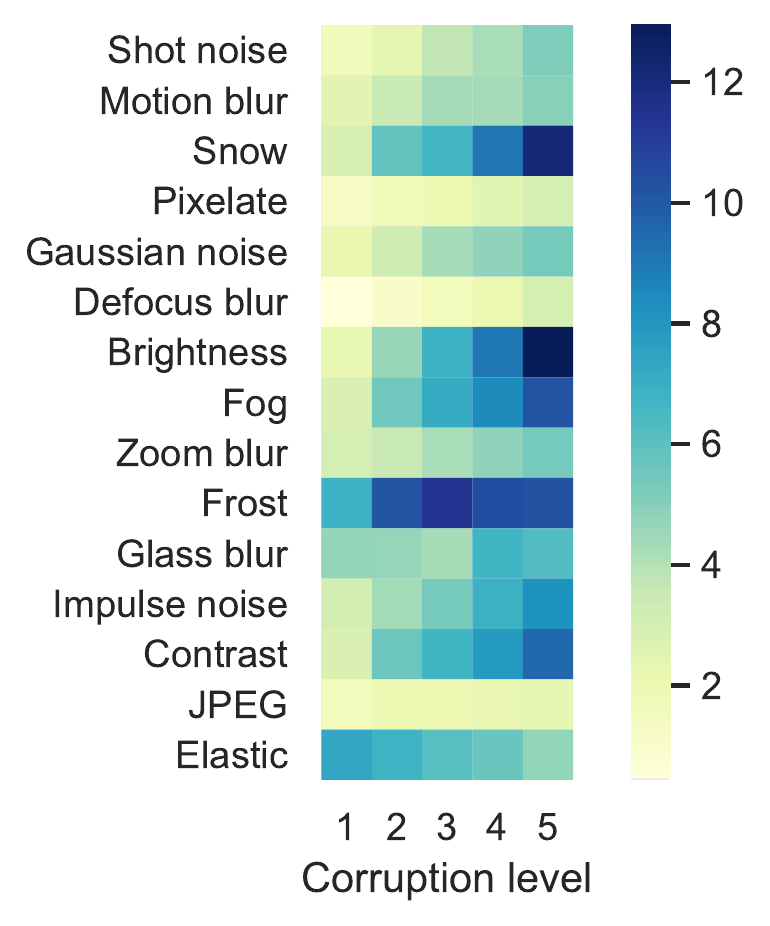}
			\includegraphics[width=0.50\columnwidth]{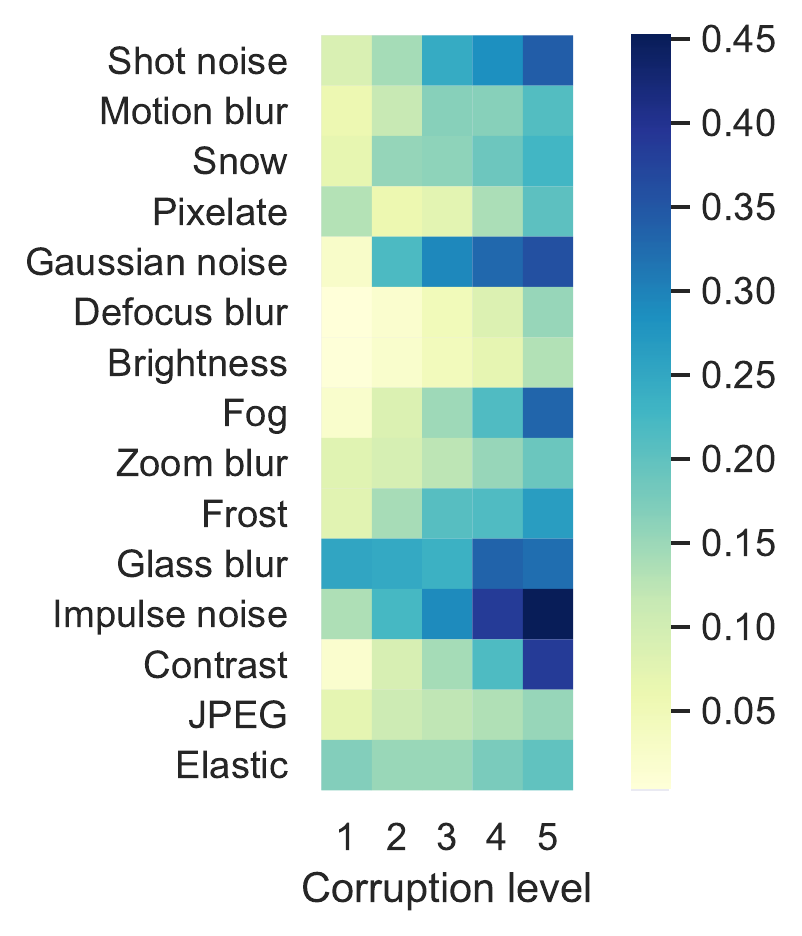}
			\caption{Average $\l_2$ and LPIPS distance for different common corruptions from CIFAR-10-C.}
			\label{fig:heatmap_distances_corruptions}
		\end{minipage}
	\end{figure}

	To investigate this behavior, we perform the following experiment in Fig.~\ref{fig:sigma_overfitting}. For random $1000$ test set images from CIFAR-10, we evaluate the loss with additive Gaussian noise of $\sigma \in [0, 0.1]$ 
	and average the loss function over both images and perturbations for (1) a standard model, (2) a model trained with Gaussian augmentation with $\sigma=0.05$ where all 100\% training samples are augmented, (3) a model trained with Gaussian augmentation for $\sigma=0.1$ where only 50\% training samples are augmented, and (4) $\l_2$ adversarially trained model with $\varepsilon=0.1$. We notice that the loss function for 100\% Gaussian augmentation is minimal at $\sigma$ which is only slightly less than $\sigma=0.05$ used for its training. \textit{Hence, the model has overfitted not only to the type of noise but also to its magnitude.} The loss function outside \emph{and inside} of the sphere is bigger than on its surface. However, there is a simple fix if we train with 50\% Gaussian noise in each batch, as suggested, e.g., in \cite{rusak2020simple} in contrast to \cite{ford2019adversarial}. This scheme allows to alleviate the $\sigma$-overfitting behavior and also achieve better accuracy on clean samples (93.2\% instead of 92.5\%) and, most importantly, \textit{significantly} improve on common corruptions (85.0\% instead of 80.5\%). At the same time, $\l_2$ adversarial training does not suffer from this problem and both 100\% and 50\% schemes work nearly equally well (details can be found in App.~\ref{sec:app_ablation_advprop_25_50_75_100}). We provide a further discussion on $\sigma$-overfitting in App.~\ref{sec:app_sigma_overfitting_details} together with additional experiments on ImageNet-100 where $\sigma$-overfitting has even more noticeable behavior. 
	
	\myparagraph{Local vs global $\l_p$ behavior.}
	Interestingly, adversarial training with worst-case perturbations bounded within a \textit{tiny} $\l_2$ ball leads to robustness significantly beyond this radius. Fig.~\ref{fig:heatmap_distances_corruptions} illustrates that common corruptions have an $\l_2$ norm an \textit{order of magnitude larger} than $\eps=0.1$ used for $\l_2$ adversarial training.
	This is in contrast with adversarial robustness that does not significantly extend beyond the radius used for training \citep{madry2018towards}. 
	Related to this, \cite{ford2019adversarial} argue that \textit{for Gaussian noise} improving the minimum distance to the decision boundary (e.g. via adversarial training) also leads to an improvement of the average distance. We have a similar mechanism at play for adversarial $\l_2$ perturbations and common corruptions which may explain the generalization of adversarial training to large average-case perturbations. 
	However, our setting is more complex compared to \cite{ford2019adversarial} since at the training and test time we deal with \textit{different} and \textit{diverse} types of noise.
	%

	\section{Improving adversarial training by relaxing a perceptual distance}
	\label{sec:rlat}
	As shown above, $\l_p$ adversarial training already leads to encouraging results on common corruptions. 
	Moreover, the $\l_2$ distance appears to be more suitable for adversarial training than $\l_\infty$ on both datasets as implied by Table~\ref{tab:at_helps}. 
	This observation suggests that using more advanced distances such as perceptual ones can further improve corruption robustness.

	\myparagraph{From $\l_p$ distances to LPIPS.}
	One of the main disadvantages of $\l_p$-norms is that they are very sensitive under simple transformations such as rotations or translations \citep{sharif2018suitability}. One possible solution is to consider \textit{perceptual distances}\footnote{Not necessarily distances in a strict mathematical sense that assumes a certain set of axioms to hold.} which capture these invariances better such as the \textit{learned perceptual image patch similarity} (LPIPS) distance introduced in \cite{zhang2018unreasonable} and which is based on the activations of a convolutional network. The LPIPS distance is formally defined as 
	\begin{align}
	\label{eq:lpips_def}
	\lpips(x,x')^2= \sum_{l=1}^L \alpha_l\|\phi_l(x)-\phi_l(x')\|_2^2, 
	\end{align}
	where $L$ is the depth of the network, $\phi_l$ is its feature map up to the $l$-th layer, and $\{\alpha_l\}_{l=1}^L$ are some constants that weigh the contributions of the $\l_2$ distances between activations. There are two crucial elements in LPIPS: the learned network and learned coefficients $\{\alpha_l\}_{l=1}^L$. \cite{zhang2018unreasonable} propose to take a network pre-trained on ImageNet and learn coefficients on their collected dataset of human judgemenets about which images are closer to each other.
	Both \cite{zhang2018unreasonable} and \cite{laidlaw2021perceptual} argue about better suitability of LPIPS to measure image similarity. In App.~\ref{sec:app_suitability_of_lpips} we analyse the suitability of LPIPS over $\l_2$ specifically on the images from CIFAR-10-C with a detailed breakdown over corruption types. In particular, we show that the LPIPS distance is better correlated with the error rate of the network, and the increase over severity levels is more monotonic compared to $\l_2$ as can be also seen in Fig.~\ref{fig:heatmap_distances_corruptions}.

	\myparagraph{LPIPS adversarial training.}
	In view of the positive features of LPIPS, adversarial training using LPIPS appears to be a promising approach to improve the performance on common corruptions.
	The worst-case loss problem considered in~\eqref{eq:rob_opt_general} using the LPIPS distance can be formulated as:
	\begin{align}
	\label{eq:lpips_at_obj}    
	\max_{\delta} \l(x+\delta, y; \theta) \quad \text{s.t.} \quad \lpips(x,x+\delta)\leq \varepsilon.
	\end{align}
	However, this optimization problem is challenging since $\lpips$ is itself defined by a neural network, and the projection onto the LPIPS-ball---as required when using PGD to solve~\eqref{eq:lpips_at_obj}---does not admit a closed-form expression. This problem was considered in \cite{laidlaw2021perceptual} who propose two approximate attacks: the Perceptual Projected Gradient Descent (PPGD) and the Lagrangian Perceptual Attack (LPA). We discuss their approach in more detail in App.~\ref{sec:app_lpips_relaxation} but emphasize that they either need to perform an approximate projection which is computationally expensive or come up with some scheme for tuning the Lagrange multiplier $\lambda$ in the Lagrangian formulation. Furthermore, they suggest in both cases to use 10-step iterative attacks for approximate LPIPS adversarial training which limits the scalability of the method to large datasets such as ImageNet.

	\myparagraph{Relaxed LPIPS adversarial training.}
	We propose here a relaxation of the LPIPS adversarial objective~\eqref{eq:lpips_at_obj}. For the simplicity of presentation, let us start by assuming that the LPIPS distance is defined using a \textit{single} intermediate layer of the network, i.e. $\lpips(x,x') = \| \phi(x) -\phi(x')\|_2$. Then we can write a neural network $f$ as the composition of the feature map $\phi$ and the remaining part of the network $f(x)=h(\phi(x))$. 
	The LPIPS adversarial objective~\eqref{eq:lpips_at_obj} in this notation becomes 
	\begin{align*}
	\max_{\delta} \l(h(\phi(x+\delta))) \quad \text{s.t.} \quad \| \phi(x)-\phi(x+\delta) \|_2\leq \varepsilon.
	\end{align*}
	We first introduce the slack variable $\tilde \delta=\phi(x)-\phi(x+\delta)$ which allows us to rewrite the objective as 
	\begin{align*}
	\max_{\delta, \tilde \delta} \l(h(\phi(x)+\tilde \delta)) \ \ \text{s.t.} \ \ \| \tilde \delta \|_2\leq \varepsilon,  \ \  \tilde \delta=\phi(x)-\phi(x+\delta).
	\end{align*}
	Then we perform the key step: we omit the constraint on the slack variable and obtain the following relaxation
	\begin{align}
	\label{eq:lpips_relax}
	\max_{\tilde \delta} \l(h(\phi(x)+\tilde \delta)) \quad \text{s.t.} \quad \| \tilde \delta \|_2\leq \varepsilon,
	\end{align}
	i.e. we lift the requirement that there should exist a $\delta$ in the \textit{input} space that corresponds to the layerwise perturbation $\tilde \delta$. 
	
	A similar relaxation can be derived when the LPIPS distance is defined using multiple layers (see App.~\ref{sec:app_lpips_relaxation}):
	\begin{align}
	\label{eq:lpips_relax_multilayer}
	&\max_{\tilde \delta^{(1)}, \dots, \tilde \delta^{(L)}} \quad \l(g_L(\dots g_1(x+\tilde\delta^{(1)}) \dots + \tilde\delta^{(L)})) \\ 
	&\text{ s.t. } \ \ \ \| \tilde \delta^{(l)} \|_2\leq \varepsilon_l \ \ \forall l \in \mathcal{L}_{LPIPS}, \ \ \ \tilde \delta^{(l)} = 0 \ \ \forall l \not\in \mathcal{L}_{LPIPS}, \nonumber
	\end{align}
	where the network is written under its compositional form $f = g_L \circ \cdots \circ g_1$, $\mathcal{L}_{LPIPS}$ is the set of layer indices used in LPIPS and $\eps_l$ denotes the $\l_2$ bound imposed at the $l$-th layer. 
	We denote this relaxation 
	as \textit{relaxed LPIPS adversarial training} (RLAT) and solve it efficiently using a single-iteration adversarial attack similar to FGM. 
	We emphasize that the projection of each $\tilde \delta^{(l)}$ onto the corresponding $\l_2$ balls is computationally cheap to perform, unlike the LPIPS projection.
	
	Since we perform relaxation and \textit{train} the network which is also used to compute LPIPS, the exact layerwise coefficients $\alpha_l$ from the original LPIPS \cite{zhang2018unreasonable} are no longer applicable and cannot be used to set the layerwise bounds $\eps_l$. Therefore, we set our own values of $\eps_l$ which we specify in App.~\ref{sec:app_lpips_relaxation} together with detailed derivations of RLAT, its precise algorithm and other implementation details.
	Finally, we remark that related layerwise adversarial training methods have been proposed before \citep{stutz2019disentangling,volpi2018generalizing,wei2020improved}.
	However, viewing layerwise adversarial training as an efficient relaxation of LPIPS adversarial training is novel, as well as applying these methods for general robustness such as common corruptions.

	\section{Experimental evaluation of RLAT}
	\label{sec:main_exps}
	Here we first show that RLAT indeed substantially improves the LPIPS robustness. Second, we compare RLAT to other established methods and show that it consistently leads to improved accuracy and calibration on common corruptions.
	
	\begin{wrapfigure}{r}{0.43\textwidth}
		\vspace{-5mm}
		\begin{center}
			\includegraphics[width=0.43\columnwidth]{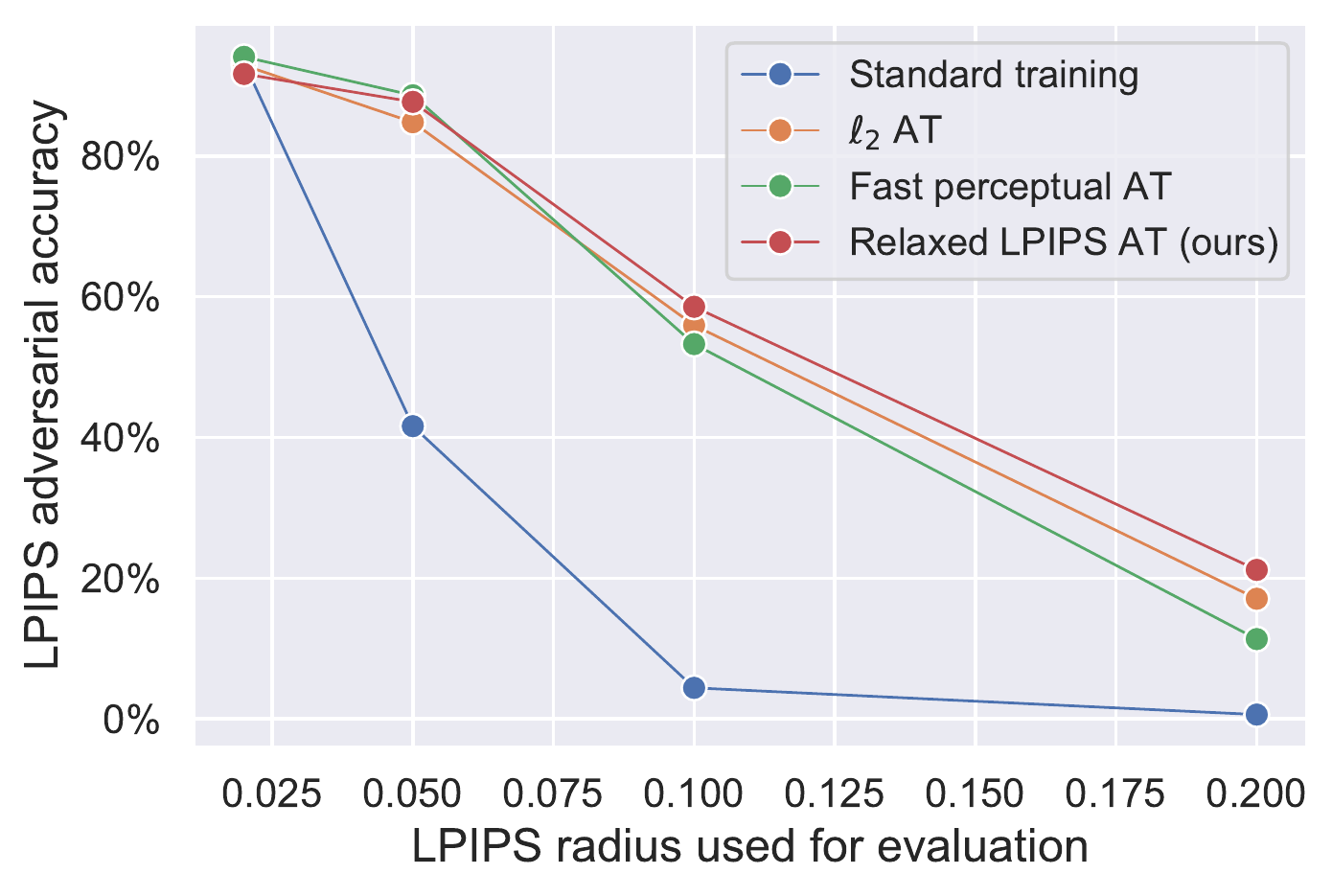}
			\vspace{-5mm}
			\caption{LPIPS adversarial robustness of different training schemes on CIFAR-10.}
			\label{fig:lpips_robustness}
		\end{center}
		\vspace{-7mm}
	\end{wrapfigure}
	\myparagraph{LPIPS robustness of RLAT.}
	We use the Lagrangian Perceptual Attack attack developed in \cite{laidlaw2021perceptual} to estimate the LPIPS adversarial accuracy under different LPIPS radii and plot results in Fig.~\ref{fig:lpips_robustness} on CIFAR-10. We use standard, $\l_2$ adversarial training (AT), Fast PAT, and RLAT models with their main hyperparameters selected to perform best on common corruptions.\footnote{We note that \cite{laidlaw2021perceptual} focus on robustness to unseen adversarial examples that involve a \textit{worst-case} optimization process, while we focus on unseen \textit{average-case} common corruptions. This is the reason why the optimal perturbation radii that we consider are noticeably smaller than in their paper.} We observe that 
	RLAT
	indeed substantially improves LPIPS robustness, even more than other approaches such as $\l_2$ AT and Fast PAT. This gives further evidence that both $\l_2$ and RLAT training \textit{do not suffer from catastrophic overfitting}, even though trained with one-step perturbations similar to FGSM. We provide a similar evaluation for $\l_2$ robustness in App.~\ref{sec:app_lpips_relaxation} (Fig.~\ref{fig:l2_robustness}).

	\myparagraph{Main experimental setup.}
	We compare the results for RLAT with additional baselines: $\l_2$~and $\l_\infty$~adversarial training (with $100\%$ adversarial samples per batch), Gaussian augmentation (with both $50\%$ and $100\%$ augmentations per batch), AdvProp \citep{xie2020adversarial}, Fast PAT~\citep{laidlaw2021perceptual}, and also four data augmentation approaches: DeepAugment \citep{hendrycks2020many}, AugMix \citep{hendrycks2019augmix}, adversarial noise training (ANT) \citep{rusak2020simple}, and Stylized ImageNet (SIN) \citep{geirhos2018imagenettrained}. 
	We use AugMix method additionally with the Jensen-Shannon regularization term as proposed in \cite{hendrycks2019augmix}.
	We train all methods from random initialization except ANT where we follow the scheme of \cite{rusak2020simple}. 
	All comparisons between methods are performed with a grid search over their main hyperparameters (reported in App.~\ref{sec:app_exp_details}) such as $\sigma$ in Gaussian augmentation or $\epsilon$ in adversarial training which we perform on the main 15 corruptions from CIFAR-10-C / ImageNet-C. In App.~\ref{sec:add_exps_app} we further verify that selecting the main hyperparameters on validation corruptions leads to the same results. 
	For Fast PAT on CIFAR-10, we do a grid search over their parameter $\eps$, but on ImageNet-100 we report the results based on the models provided by the authors due to limited computational resources.
	To assess calibration, we report the expected calibration error (ECE) (see App.~\ref{sec:add_exps_app} for ECE with temperature rescaling \cite{guo2017calibration}).
	More details can be found in our repository \url{https://github.com/tml-epfl/adv-training-corruptions}.

	Since the main goal of the common corruption benchmark \citep{hendrycks2019benchmarking} is to show the model’s behavior on \textit{unseen} corruptions, we do not use overlapping augmentations in training (see App.~\ref{sec:app_exp_details}). 
	The only exception is Gaussian augmentation which we mark in gray in Table~\ref{tab:main_cifar10_imagenet100} following \citep{rusak2020simple} since it belongs to common corruptions. 
	We note that removing only Gaussian noise from evaluation is not sufficient, because other noises can be affected as well by training with Gaussian augmentation. 
	Thus, the results of 100\% and 50\% Gaussian augmentation are shown only for illustrative purposes suggesting that adversarial training with no prior knowledge about the corruptions can obtain almost the same results as direct augmentation.

	\begin{table}[t!]
		\centering
		\small
		\setlength{\tabcolsep}{4.0pt}
		\begin{tabular}{@{}lccc}
			& \textbf{Standard} & \textbf{Corruption} & \textbf{Corruption} \\
			\textbf{Training} & \textbf{accuracy} & \textbf{accuracy}   & \textbf{calibr. error}  \\
			\midrule
			\\
			& \multicolumn{3}{c}{\textbf{CIFAR-10}}  \\
			\cmidrule(lr){2-4}
			Standard & 95.1\% & 74.6\% & 16.6\% \\
			100\% Gaussian & 92.5\% & \color{gray} 80.5\% & \color{gray} 13.2\%  \\ 
			50\%  Gaussian & 93.2\% & \color{gray} 85.0\% & \color{gray} 9.1\%   \\ 
			Fast PAT & 93.4\% & 80.6\% & 12.0\%   \\
			AdvProp  & 94.7\% & 82.9\% & 10.1\%     \\
			$\l_\infty$ adversarial  & 93.3\% & 82.7\% & 10.8\% \\
			$\l_2$ adversarial & 93.6\% & 83.4\% & 10.5\% \\
			RLAT & 93.1\% & \textbf{84.1\%} & \textbf{9.9\%} \\
			\midrule
			DeepAugment & 94.1\% & 85.3\% & 8.7\% \\
			DeepAugment + RLAT & 93.6\% & \textbf{87.8\%} & \textbf{6.1\%} \\
			\midrule
			AugMix & 95.0\% & 86.6\% & 6.9\%  \\ 
			AugMix + RLAT & 94.8\% & \textbf{88.5\%} & \textbf{4.5\%}  \\ 
			\midrule
			AugMix + JSD & 95.0\% & 88.6\% & 6.5\%  \\
			AugMix + JSD + RLAT & 94.8\% & \textbf{89.6\%}  & \textbf{5.4\%}  \\  
			\\
			& \multicolumn{3}{c}{\textbf{ImageNet-100}}  \\
			\cmidrule(lr){2-4}
			Standard & 86.6\% & 47.5\% & 10.0\% \\
			100\% Gaussian & 86.4\% & \color{gray} 46.7\% & \color{gray} 11.7\% \\ 
			50\%  Gaussian & 83.8\% & \color{gray} 55.2\% & \color{gray} 6.1\% \\ 
			Fast PAT  & 71.5\% & 45.2\% & 8.0\% \\
			$\l_\infty$ adversarial &  86.5\% & 47.7\% & 12.4\% \\
			$\l_2$ adversarial &  86.3\% & 48.4\% & 9.4\% \\
			RLAT & 86.5\% & \textbf{48.8\%} & \textbf{9.1\%} \\
			\midrule
			AugMix & 86.7\% & 52.3\% & 7.5\% \\ 
			AugMix + RLAT & 86.8\% & \textbf{54.8\%} & \textbf{4.7\%} \\ 
			\midrule
			AugMix + JSD &   88.4\% & 59.3\% & 1.9\% \\
			AugMix + JSD + RLAT & 87.1\% & \textbf{61.1\%} & \textbf{1.8\%} \\  
			
			\midrule
			SIN & 86.6\% & 53.7\% & 6.7\% \\  
			SIN + RLAT & 86.5\% & \textbf{54.3\%} & \textbf{6.0\%} \\   
			\midrule
			ANT\textsuperscript{3x3}  & 85.9\% & 57.7\% & 5.1\% \\ 
			ANT\textsuperscript{3x3} + RLAT  & 85.3\% & \textbf{58.3\%} & \textbf{4.4\%} \\
		\end{tabular}
		\caption{Accuracy and calibration of ResNet-18 models trained on CIFAR-10 and ImageNet-100. Gray-colored numbers correspond to methods partially trained with the corruptions from CIFAR-10-C and ImageNet-100-C.}
		\label{tab:main_cifar10_imagenet100}
	\end{table}
	
	\myparagraph{Main experimental results.}
	We show the main experimental results on CIFAR-10-C and ImageNet-100-C in Table~\ref{tab:main_cifar10_imagenet100}. 
	First of all, we observe that $\l_p$ adversarial training is a strong baseline on common corruptions on both datasets with a larger gain on CIFAR-10-C. 
	Using our proposed relaxed LPIPS adversarial training further improves the corruption accuracy on both datasets: from 74.6\% to 84.1\% on CIFAR-10-C and from 47.5\% to 48.8\% compared to standard models. Moreover, RLAT also improves calibration compared to the standard model: from 16.6\% to 9.9\% ECE on CIFAR-10-C and from 10.0\% to 9.1\% ECE on ImageNet-100-C.
	We also observe that 100\% Gaussian augmentation even deteriorates the performance on ImageNet-100-C
	while 50\% Gaussian augmentation significantly improves the average accuracy which is consistent with \citet{rusak2020simple}.
	
	We observe that RLAT can be successfully combined with existing data augmentations, leading to better accuracy and calibration. E.g.,
	adding RLAT on top of DeepAugment helps to improve the CIFAR-10-C accuracy from 85.3\% to 87.8\%.
	Combining RLAT with the AugMix augmentation improves the corruption accuracy from 86.6 \% to 88.5\% on CIFAR-10-C and on ImageNet-100-C from 52.3\% to 54.8\%. Combining SIN and ANT\textsuperscript{3x3} improves the accuracy on ImageNet-100-C from 53.7\% to 54.3\% and from 57.7\% to 58.3\%, respectively. 
	Moreover, we see that RLAT consistently improves ECE in all settings, and we refer to App.~\ref{sec:add_exps_app} for ECE with temperature rescaling 
	which qualitatively shows the same behavior.
	Thus, RLAT is not only a helpful technique on its own but can also benefit from advanced data augmentations.

	\myparagraph{Runtime of RLAT.}
	RLAT achieves a significant speed-up over Fast PAT: 1.8 hours vs 9.4 hours of wallclock time on CIFAR-10.
	The runtime of RLAT is not much higher than the runtime of $\l_2/\l_\infty$ adversarial training (1.8 hours vs 1.3 hours on CIFAR-10) and standard training (0.8 hours on CIFAR-10).
	On ImageNet-100, RLAT takes 6.2 hours on a single V100 GPU which can be compared to 120 hours on 4 Nvidia RTX 2080 Ti GPUs for Fast PAT (although Fast PAT uses a larger network, ResNet-50 instead of ResNet-18).
	%
	We report a full runtime comparison in App.~\ref{sec:app_exp_details}. 
	

	\myparagraph{Additional experiments.}
	We provide additional experimental results in the Appendix. 
	In App.~\ref{sec:app_distr_shifts}, we evaluate the performance of the models from Table~\ref{tab:main_cifar10_imagenet100} on ImageNet-A, ImageNet-R, and Stylized ImageNet to better understand how well the improvements on common corruptions transfer to other distribution shifts. In App.~\ref{sec:add_exps_app}, we provide more detailed experimental results such as those presented in Table~\ref{tab:main_cifar10_imagenet100} but with breakdowns over different corruptions and severities. We also present results for larger network architectures and for AugMix combined with $\l_p$ adversarial training in App.~\ref{sec:add_exps_app}, as well as results of RLAT over multiple random seeds.

	\section{Conclusions and future work}
	Our findings suggest that adversarial training can be successfully used to improve accuracy and calibration on common image corruptions. 
	Even simple $\l_p$ adversarial training can serve as a strong baseline if the optimal perturbation radius $\eps$ is carefully chosen for the given problem.
	More advanced adversarial training schemes involve perceptual distances, such as LPIPS, and we provide a relaxation of LPIPS adversarial training with an efficient single-step procedure for adversarial training. We observe that the developed relaxation substantially improves the LPIPS robustness and can be successfully combined with existing data augmentations.
	We hope that the developed relaxed LPIPS adversarial training would be of interest also for other domains such as natural language processing where robustness to commonly occurring corruptions (e.g., typos) is an important task.

	\section*{Acknowledgments}
	We thank the authors of \cite{laidlaw2021perceptual} for providing the code of their method. 
	We also thank Francesco Croce and Vikash Sehwag for many fruitful discussions.

	\small
	\bibliography{literature}

\begin{thebibliography}{56}
\providecommand{\natexlab}[1]{#1}
\providecommand{\url}[1]{\texttt{#1}}
\expandafter\ifx\csname urlstyle\endcsname\relax
  \providecommand{\doi}[1]{doi: #1}\else
  \providecommand{\doi}{doi: \begingroup \urlstyle{rm}\Url}\fi

\bibitem[Andriushchenko \& Flammarion(2020)Andriushchenko and
  Flammarion]{andriushchenko2020understanding}
Andriushchenko, M. and Flammarion, N.
\newblock Understanding and improving fast adversarial training.
\newblock In \emph{NeurIPS}, 2020.

\bibitem[Azulay \& Weiss(2019)Azulay and Weiss]{azulay2019deep}
Azulay, A. and Weiss, Y.
\newblock Why do deep convolutional networks generalize so poorly to small
  image transformations?
\newblock \emph{JMLR}, 20\penalty0 (184):\penalty0 1--25, 2019.

\bibitem[Bishop(1995)]{bishop1995training}
Bishop, C.~M.
\newblock Training with noise is equivalent to {T}ikhonov regularization.
\newblock \emph{Neural Computation}, 7\penalty0 (1):\penalty0 108–116,
  January 1995.

\bibitem[Croce \& Hein(2020)Croce and Hein]{croce2020reliable}
Croce, F. and Hein, M.
\newblock Reliable evaluation of adversarial robustness with an ensemble of
  diverse parameter-free attacks.
\newblock In \emph{ICML}, 2020.

\bibitem[Croce et~al.(2020)Croce, Andriushchenko, Sehwag, Flammarion, Chiang,
  Mittal, and Hein]{croce2020robustbench}
Croce, F., Andriushchenko, M., Sehwag, V., Flammarion, N., Chiang, M., Mittal,
  P., and Hein, M.
\newblock Robustbench: a standardized adversarial robustness benchmark.
\newblock \emph{arXiv preprint arXiv:2010.09670}, 2020.

\bibitem[Cubuk et~al.(2019)Cubuk, Zoph, Mane, Vasudevan, and
  Le]{cubuk2018autoaugment}
Cubuk, E.~D., Zoph, B., Mane, D., Vasudevan, V., and Le, Q.~V.
\newblock Autoaugment: Learning augmentation policies from data.
\newblock In \emph{CVPR}, 2019.

\bibitem[DeVries \& Taylor(2017)DeVries and Taylor]{devries2017improved}
DeVries, T. and Taylor, G.~W.
\newblock Improved regularization of convolutional neural networks with cutout.
\newblock \emph{arXiv preprint arXiv:1708.04552}, 2017.

\bibitem[Dodge \& Karam(2017)Dodge and Karam]{dodge2017study}
Dodge, S. and Karam, L.
\newblock A study and comparison of human and deep learning recognition
  performance under visual distortions.
\newblock In \emph{ICCCN}, 2017.

\bibitem[Drucker \& LeCun(1992)Drucker and LeCun]{drucker1992improving}
Drucker, H. and LeCun, Y.
\newblock Improving generalization performance using double backpropagation.
\newblock \emph{IEEE Transactions on Neural Networks}, 1992.

\bibitem[Engstrom et~al.(2019)Engstrom, Ilyas, Salman, Santurkar, and
  Tsipras]{engstrom2019robustness}
Engstrom, L., Ilyas, A., Salman, H., Santurkar, S., and Tsipras, D.
\newblock Robustness (python library), 2019.
\newblock URL \url{https://github.com/MadryLab/robustness}.

\bibitem[Ford et~al.(2019)Ford, Gilmer, Carlini, and
  Cubuk]{ford2019adversarial}
Ford, N., Gilmer, J., Carlini, N., and Cubuk, D.
\newblock Adversarial examples are a natural consequence of test error in
  noise.
\newblock In \emph{ICML}, 2019.

\bibitem[Geirhos et~al.(2018)Geirhos, Temme, Rauber, Sch{\"u}tt, Bethge, and
  Wichmann]{geirhos2018generalisation}
Geirhos, R., Temme, C. R.~M., Rauber, J., Sch{\"u}tt, H.~H., Bethge, M., and
  Wichmann, F.~A.
\newblock Generalisation in humans and deep neural networks.
\newblock In \emph{NeurIPS}, 2018.

\bibitem[Geirhos et~al.(2019)Geirhos, Rubisch, Michaelis, Bethge, Wichmann, and
  Brendel]{geirhos2018imagenettrained}
Geirhos, R., Rubisch, P., Michaelis, C., Bethge, M., Wichmann, F.~A., and
  Brendel, W.
\newblock Imagenet-trained {CNN}s are biased towards texture; increasing shape
  bias improves accuracy and robustness.
\newblock \emph{ICLR}, 2019.

\bibitem[Goodfellow et~al.(2015)Goodfellow, Shlens, and
  Szegedy]{goodfellow2014explaining}
Goodfellow, I.~J., Shlens, J., and Szegedy, C.
\newblock Explaining and harnessing adversarial examples.
\newblock \emph{ICLR}, 2015.

\bibitem[Guo et~al.(2017)Guo, Pleiss, Sun, and Weinberger]{guo2017calibration}
Guo, C., Pleiss, G., Sun, Y., and Weinberger, K.~Q.
\newblock On calibration of modern neural networks.
\newblock In \emph{International Conference on Machine Learning}, pp.\
  1321--1330. PMLR, 2017.

\bibitem[He et~al.(2016)He, Zhang, Ren, and Sun]{he2016identity}
He, K., Zhang, X., Ren, S., and Sun, J.
\newblock Identity mappings in deep residual networks.
\newblock \emph{ECCV}, 2016.

\bibitem[Hendrycks \& Dietterich(2019)Hendrycks and
  Dietterich]{hendrycks2019benchmarking}
Hendrycks, D. and Dietterich, T.
\newblock Benchmarking neural network robustness to common corruptions and
  perturbations.
\newblock In \emph{ICLR}, 2019.

\bibitem[Hendrycks et~al.(2019{\natexlab{a}})Hendrycks, Lee, and
  Mazeika]{hendrycks2019using}
Hendrycks, D., Lee, K., and Mazeika, M.
\newblock Using pre-training can improve model robustness and uncertainty.
\newblock In \emph{ICML}, 2019{\natexlab{a}}.

\bibitem[Hendrycks et~al.(2019{\natexlab{b}})Hendrycks, Mu, Cubuk, Zoph,
  Gilmer, and Lakshminarayanan]{hendrycks2019augmix}
Hendrycks, D., Mu, N., Cubuk, E.~D., Zoph, B., Gilmer, J., and
  Lakshminarayanan, B.
\newblock Augmix: A simple data processing method to improve robustness and
  uncertainty.
\newblock In \emph{ICLR}, 2019{\natexlab{b}}.

\bibitem[Hendrycks et~al.(2019{\natexlab{c}})Hendrycks, Zhao, Basart,
  Steinhardt, and Song]{hendrycks2019natural}
Hendrycks, D., Zhao, K., Basart, S., Steinhardt, J., and Song, D.
\newblock Natural adversarial examples.
\newblock \emph{arXiv preprint arXiv:1907.07174}, 2019{\natexlab{c}}.

\bibitem[Hendrycks et~al.(2021)Hendrycks, Basart, Mu, Kadavath, Wang, Dorundo,
  Desai, Zhu, Parajuli, Guo, et~al.]{hendrycks2020many}
Hendrycks, D., Basart, S., Mu, N., Kadavath, S., Wang, F., Dorundo, E., Desai,
  R., Zhu, T., Parajuli, S., Guo, M., et~al.
\newblock The many faces of robustness: A critical analysis of
  out-of-distribution generalization.
\newblock In \emph{ICCV}, 2021.

\bibitem[Huang et~al.(2020)Huang, Menkovski, Pei, and
  Pechenizkiy]{huang2020bridging}
Huang, T., Menkovski, V., Pei, Y., and Pechenizkiy, M.
\newblock Bridging the performance gap between fgsm and pgd adversarial
  training.
\newblock \emph{arXiv preprint arXiv:2011.05157}, 2020.

\bibitem[Ilyas et~al.(2019)Ilyas, Santurkar, Tsipras, Engstrom, Tran, and
  Madry]{ilyas2019adversarial}
Ilyas, A., Santurkar, S., Tsipras, D., Engstrom, L., Tran, B., and Madry, A.
\newblock Adversarial examples are not bugs, they are features.
\newblock In \emph{NeurIPS}, 2019.

\bibitem[Kang et~al.(2019)Kang, Sun, Brown, Hendrycks, and
  Steinhardt]{kang2019transfer}
Kang, D., Sun, Y., Brown, T., Hendrycks, D., and Steinhardt, J.
\newblock Transfer of adversarial robustness between perturbation types.
\newblock \emph{arXiv preprint arXiv:1905.01034}, 2019.

\bibitem[Krizhevsky \& Hinton(2009)Krizhevsky and
  Hinton]{krizhevsky2009learning}
Krizhevsky, A. and Hinton, G.
\newblock Learning multiple layers of features from tiny images.
\newblock \emph{Technical Report}, 2009.

\bibitem[Krizhevsky et~al.(2012)Krizhevsky, Sutskever, and
  Hinton]{krizhevsky2012imagenet}
Krizhevsky, A., Sutskever, I., and Hinton, G.
\newblock Imagenet classification with deep convolutional neural networks.
\newblock In \emph{NeurIPS}, 2012.

\bibitem[Laidlaw et~al.(2021)Laidlaw, Singla, and Feizi]{laidlaw2021perceptual}
Laidlaw, C., Singla, S., and Feizi, S.
\newblock Perceptual adversarial robustness: Generalizable defenses against
  unforeseen threat models.
\newblock In \emph{ICLR}, 2021.

\bibitem[Madry et~al.(2018)Madry, Makelov, Schmidt, Tsipras, and
  Vladu]{madry2018towards}
Madry, A., Makelov, A., Schmidt, L., Tsipras, D., and Vladu, A.
\newblock Towards deep learning models resistant to adversarial attacks.
\newblock In \emph{ICLR}, 2018.

\bibitem[Mintun et~al.(2021)Mintun, Kirillov, and Xie]{mintun2021interaction}
Mintun, E., Kirillov, A., and Xie, S.
\newblock On interaction between augmentations and corruptions in natural
  corruption robustness.
\newblock \emph{arXiv preprint arXiv:2102.11273}, 2021.

\bibitem[Moosavi-Dezfooli et~al.(2019)Moosavi-Dezfooli, Fawzi, Uesato, and
  Frossard]{moosavi2019robustness}
Moosavi-Dezfooli, S.-M., Fawzi, A., Uesato, J., and Frossard, P.
\newblock Robustness via curvature regularization, and vice versa.
\newblock In \emph{CVPR}, 2019.

\bibitem[Nandy et~al.(2021)Nandy, Saha, Hsu, Lee, and
  Zhu]{nandy2021adversarially}
Nandy, J., Saha, S., Hsu, W., Lee, M.~L., and Zhu, X.~X.
\newblock Adversarially trained models with test-time covariate shift
  adaptation.
\newblock \emph{arXiv}, 2021.

\bibitem[Ortiz-Jimenez et~al.(2020)Ortiz-Jimenez, Modas, Moosavi-Dezfooli, and
  Frossard]{ortiz2020hold}
Ortiz-Jimenez, G., Modas, A., Moosavi-Dezfooli, S.-M., and Frossard, P.
\newblock Hold me tight! influence of discriminative features on deep network
  boundaries.
\newblock In \emph{NeurIPS}, 2020.

\bibitem[Ovadia et~al.(2019)Ovadia, Fertig, Ren, Nado, Sculley, Nowozin,
  Dillon, Lakshminarayanan, and Snoek]{ovadia2019can}
Ovadia, Y., Fertig, E., Ren, J., Nado, Z., Sculley, D., Nowozin, S., Dillon,
  J.~V., Lakshminarayanan, B., and Snoek, J.
\newblock Can you trust your model's uncertainty? evaluating predictive
  uncertainty under dataset shift.
\newblock \emph{arXiv preprint arXiv:1906.02530}, 2019.

\bibitem[Radford et~al.(2021)Radford, Kim, Hallacy, Ramesh, Goh, Agarwal,
  Sastry, Askell, Mishkin, Clark, et~al.]{radford2021learning}
Radford, A., Kim, J.~W., Hallacy, C., Ramesh, A., Goh, G., Agarwal, S., Sastry,
  G., Askell, A., Mishkin, P., Clark, J., et~al.
\newblock Learning transferable visual models from natural language
  supervision.
\newblock 2021.

\bibitem[Rusak et~al.(2020)Rusak, Schott, Zimmermann, Bitterwolf, Bringmann,
  Bethge, and Brendel]{rusak2020simple}
Rusak, E., Schott, L., Zimmermann, R.~S., Bitterwolf, J., Bringmann, O.,
  Bethge, M., and Brendel, W.
\newblock A simple way to make neural networks robust against diverse image
  corruptions.
\newblock In \emph{ECCV}, 2020.

\bibitem[Russakovsky et~al.(2015)Russakovsky, Deng, Su, Krause, Satheesh, Ma,
  Huang, Karpathy, Khosla, Bernstein, et~al.]{russakovsky2015imagenet}
Russakovsky, O., Deng, J., Su, H., Krause, J., Satheesh, S., Ma, S., Huang, Z.,
  Karpathy, A., Khosla, A., Bernstein, M., et~al.
\newblock Imagenet large scale visual recognition challenge.
\newblock \emph{IJCV}, 2015.

\bibitem[Schneider et~al.(2020)Schneider, Rusak, Eck, Bringmann, Brendel, and
  Bethge]{schneider2020improving}
Schneider, S., Rusak, E., Eck, L., Bringmann, O., Brendel, W., and Bethge, M.
\newblock Improving robustness against common corruptions by covariate shift
  adaptation.
\newblock In \emph{NeurIPS}, 2020.

\bibitem[Shafahi et~al.(2019)Shafahi, Najibi, Ghiasi, Xu, Dickerson, Studer,
  Davis, Taylor, and Goldstein]{shafahi2019adversarial}
Shafahi, A., Najibi, M., Ghiasi, A., Xu, Z., Dickerson, J., Studer, C., Davis,
  L.~S., Taylor, G., and Goldstein, T.
\newblock Adversarial training for free!
\newblock In \emph{NeurIPS}, 2019.

\bibitem[Shafahi et~al.(2020)Shafahi, Najibi, Xu, Dickerson, Davis, and
  Goldstein]{shafahi2020universal}
Shafahi, A., Najibi, M., Xu, Z., Dickerson, J., Davis, L.~S., and Goldstein, T.
\newblock Universal adversarial training.
\newblock In \emph{AAAI}, 2020.

\bibitem[Shaham et~al.(2018)Shaham, Yamada, and
  Negahban]{shaham2015understanding}
Shaham, U., Yamada, Y., and Negahban, S.
\newblock Understanding adversarial training: Increasing local stability of
  supervised models through robust optimization.
\newblock \emph{Neurocomputing}, 2018.

\bibitem[Sharif et~al.(2018)Sharif, Bauer, and Reiter]{sharif2018suitability}
Sharif, M., Bauer, L., and Reiter, M.~K.
\newblock On the suitability of lp-norms for creating and preventing
  adversarial examples.
\newblock In \emph{CVPR Workshops}, 2018.

\bibitem[Simon-Gabriel et~al.(2019)Simon-Gabriel, Ollivier, Bottou,
  Sch{\"o}lkopf, and Lopez-Paz]{simon2019first}
Simon-Gabriel, C.-J., Ollivier, Y., Bottou, L., Sch{\"o}lkopf, B., and
  Lopez-Paz, D.
\newblock First-order adversarial vulnerability of neural networks and input
  dimension.
\newblock \emph{ICML}, 2019.

\bibitem[Stutz et~al.(2019)Stutz, Hein, and Schiele]{stutz2019disentangling}
Stutz, D., Hein, M., and Schiele, B.
\newblock Disentangling adversarial robustness and generalization.
\newblock In \emph{CVPR}, 2019.

\bibitem[Szegedy et~al.(2014)Szegedy, Zaremba, Sutskever, Bruna, Erhan,
  Goodfellow, and Fergus]{szegedy2013intriguing}
Szegedy, C., Zaremba, W., Sutskever, I., Bruna, J., Erhan, D., Goodfellow, I.,
  and Fergus, R.
\newblock Intriguing properties of neural networks.
\newblock In \emph{ICLR}, 2014.

\bibitem[Taori et~al.(2020)Taori, Dave, Shankar, Carlini, Recht, and
  Schmidt]{taori2020measuring}
Taori, R., Dave, A., Shankar, V., Carlini, N., Recht, B., and Schmidt, L.
\newblock Measuring robustness to natural distribution shifts in image
  classification.
\newblock \emph{NeurIPS}, 2020.

\bibitem[Tramèr et~al.(2018)Tramèr, Kurakin, Papernot, Goodfellow, Boneh, and
  McDaniel]{tramer2018ensemble}
Tramèr, F., Kurakin, A., Papernot, N., Goodfellow, I., Boneh, D., and
  McDaniel, P.
\newblock Ensemble adversarial training: Attacks and defenses.
\newblock In \emph{ICLR}, 2018.

\bibitem[Tsipras et~al.(2019)Tsipras, Santurkar, Engstrom, Turner, and
  Madry]{tsipras2018robustness}
Tsipras, D., Santurkar, S., Engstrom, L., Turner, A., and Madry, A.
\newblock Robustness may be at odds with accuracy.
\newblock In \emph{ICLR}, 2019.

\bibitem[Volpi et~al.(2018)Volpi, Namkoong, Sener, Duchi, Murino, and
  Savarese]{volpi2018generalizing}
Volpi, R., Namkoong, H., Sener, O., Duchi, J.~C., Murino, V., and Savarese, S.
\newblock Generalizing to unseen domains via adversarial data augmentation.
\newblock In \emph{NeurIPS}, 2018.

\bibitem[Wei \& Ma(2020)Wei and Ma]{wei2020improved}
Wei, C. and Ma, T.
\newblock Improved sample complexities for deep neural networks and robust
  classification via an all-layer margin.
\newblock In \emph{ICLR}, 2020.

\bibitem[Wong et~al.(2020)Wong, Rice, and Kolter]{wong2020fast}
Wong, E., Rice, L., and Kolter, J.~Z.
\newblock Fast is better than free: Revisiting adversarial training.
\newblock In \emph{ICLR}, 2020.

\bibitem[Xie et~al.(2019)Xie, Wu, Maaten, Yuille, and He]{xie2019feature}
Xie, C., Wu, Y., Maaten, L. v.~d., Yuille, A.~L., and He, K.
\newblock Feature denoising for improving adversarial robustness.
\newblock In \emph{CVPR}, 2019.

\bibitem[Xie et~al.(2020)Xie, Tan, Gong, Wang, Yuille, and
  Le]{xie2020adversarial}
Xie, C., Tan, M., Gong, B., Wang, J., Yuille, A.~L., and Le, Q.~V.
\newblock Adversarial examples improve image recognition.
\newblock In \emph{CVPR}, 2020.

\bibitem[Yin et~al.(2019)Yin, Lopes, Shlens, Cubuk, and Gilmer]{yin2019fourier}
Yin, D., Lopes, R.~G., Shlens, J., Cubuk, E.~D., and Gilmer, J.
\newblock A fourier perspective on model robustness in computer vision.
\newblock In \emph{NeurIPS}, 2019.

\bibitem[Yun et~al.(2019)Yun, Han, Oh, Chun, Choe, and Yoo]{yun2019cutmix}
Yun, S., Han, D., Oh, S.~J., Chun, S., Choe, J., and Yoo, Y.
\newblock Cutmix: Regularization strategy to train strong classifiers with
  localizable features.
\newblock In \emph{ICCV}, 2019.

\bibitem[Zhang et~al.(2018{\natexlab{a}})Zhang, Cisse, Dauphin, and
  Lopez-Paz]{zhang2018mixup}
Zhang, H., Cisse, M., Dauphin, Y.~N., and Lopez-Paz, D.
\newblock mixup: Beyond empirical risk minimization.
\newblock In \emph{ICLR}, 2018{\natexlab{a}}.

\bibitem[Zhang et~al.(2018{\natexlab{b}})Zhang, Isola, Efros, Shechtman, and
  Wang]{zhang2018unreasonable}
Zhang, R., Isola, P., Efros, A.~A., Shechtman, E., and Wang, O.
\newblock The unreasonable effectiveness of deep features as a perceptual
  metric.
\newblock In \emph{CVPR}, 2018{\natexlab{b}}.

\end{thebibliography}
	\bibliographystyle{icml2021}
	\clearpage
	\normalsize

	\normalsize

	\appendix

	
	\begin{center}
		\LARGE\textbf{Appendix}
	\end{center}

	\section*{Organization of the Appendix}
	The appendix contains additional implementation details for our methods and the baselines we compare to, as well as more detailed derivations and experimental results.
	The appendix is organized as follows:
	\begin{itemize}
		\item Sec.~\ref{sec:app_exp_details}: further details on our experimental setup, hyperparameter choice, and runtime of different methods. 
		\item Sec.~\ref{sec:app_calibration}: further details on the calibration experiments and results for $\l_2$ adversarial training.
		\item Sec.~\ref{sec:app_ablation_advprop_25_50_75_100}: an ablation study for AdvProp comparing it to adversarial training with some fraction of clean images.
		\item Sec.~\ref{sec:app_sigma_overfitting_details}: further experiments related to $\sigma$-overfitting on ImageNet-100 and more detailed discussion. 
		\item Sec.~\ref{sec:app_suitability_of_lpips}: discussion on why LPIPS distance is particularly suitable for the corrupted images from CIFAR-10-C.
		\item Sec.~\ref{sec:app_lpips_relaxation}: full derivations for our Relaxed LPIPS adversarial training method, further implementation details, and evaluation of $\l_2$ robustness.
		\item Sec.~\ref{sec:app_distr_shifts}: evaluation of the performance of various models on different distribution shifts such as ImageNet-A, ImageNet-R, and Stylized ImageNet.
		\item Sec.~\ref{sec:add_exps_app}: more detailed experimental results related to Sections~\ref{sec:at_vs_gauss} and~\ref{sec:main_exps} such as breakdowns over different corruptions and severity levels.
	\end{itemize}

	\section{Experimental details}
	\label{sec:app_exp_details}
	In this section, we provide more details regarding our experimental settings, hyperparameters, evaluation metrics, and runtime of our method.
	
	\myparagraph{Dataset details.}  
	We perform experiments on two common image classification datasets: CIFAR-10 \citep{krizhevsky2009learning} which has $32\times32$ images, and ImageNet-100 \citep{russakovsky2015imagenet} with $224\times224$ images where we take each tenth class according to the WordNet ID order following \cite{laidlaw2021perceptual}. We choose ImageNet-100 instead of the full ImageNet since we have limited computation resources for performing grid searches on large-scale datasets over the main hyperparameters such as the perturbation radius for adversarial training or the standard deviation of the Gaussian noise.
	
	The CIFAR-10-C and ImageNet-C datasets that were introduced in \cite{hendrycks2019benchmarking} contain $15$ main synthetic corruptions: Gaussian noise, shot noise, impulse noise, defocus blur, glass blur, motion blur, zoom blur, snow, frost, fog, brightness, contrast, elastic, pixelation, and JPEG.
	Both datasets contain also $4$ additional corruptions (speckle noise, Gaussian blur, spatter, saturation) that are not commonly used. 
	Each corruption has 5 levels of severity. 
	We use the CIFAR-10-C and ImageNet-C images provided by \cite{hendrycks2019benchmarking}, although one could alternatively apply the corruptions in-memory (as done, e.g., in \citep{ford2019adversarial}).
	
	In addition, we make use of three more variants of ImageNet: Stylized ImageNet, ImageNet-A and ImageNet-R. Stylized ImageNet (SIN) is a variant of ImageNet which is obtained using style transfer. It has been first introduced to induce a shape bias in convolutional networks \citep{geirhos2018imagenettrained}.
	ImageNet-A \citep{hendrycks2019natural} is a test set of $7\,500$ natural but adversarially collected images with, e.g., unusual backgrounds or occlusions for 200 ImageNet classes.
	ImageNet-R \citep{hendrycks2020many} is a test set of $30\,000$ image renditions (e.g., paintings, sculptures, embroidery) for another set of 200 ImageNet classes.
	When evaluating on these datasets, we only use the classes intersecting with ImageNet-100.
	
	\begin{table*}[t]
		\centering
		\footnotesize
		\begin{tabular}{@{}lll@{}}
			& \multicolumn{2}{c}{\textbf{Dataset}} \\
			\cmidrule(lr){2-3}
			\textbf{Hyperparameter} & CIFAR-10 & ImageNet-100 \\
			\midrule
			Architecture & PreAct ResNet-18 & PreAct ResNet-18 \\
			Number of epochs       & 150 & 100 \\
			Learning rate of SGD   & 0.1 & 0.1 \\
			Epochs for learning rate decay (by $10\times$ factor) & \{50, 100\} & \{33, 66\} \\
			Momentum               & 0.9 & 0.9 \\
			Batch size             & 128 & 128 \\
			Weight decay           & 0.0005 & 0.0005 \\
		\end{tabular}
		\caption{The main hyperparameters used for CIFAR-10 and ImageNet experiments.}
		\label{tab:app_hyperparams_main}
	\end{table*}
	\begin{table*}[t]
		\centering
		\footnotesize
		\begin{tabular}{@{}ll@{}}
			\textbf{Method} & G\textbf{rid values} \\
			\midrule
			100\% Gaussian augmentation & $\sigma \in \{0.02, \textbf{0.05}, 0.08, 0.1, 0.2\}$ \\
			50\% Gaussian augmentation & $\sigma \in \{0.02, 0.05, 0.08, \textbf{0.1}, 0.2\}$ \\
			$\l_\infty$ adversarial training & $\eps \in \{0.1, 0.5, \textbf{1.0}, 2.0, 4.0, 8.0, 16.0\} / 255$ \\
			$\l_2$ adversarial training & $\eps \in \{0.01, 0.05, 0.08, \textbf{0.1}, 0.15, 0.2, 0.5, 1.0\}$ \\
			Fast PAT & $\eps \in \{0.005, 0.01, \textbf{0.02}, 0.05, 0.08\}$ \\
			AdvProp & $\eps \in \{0.5, 1.0, \textbf{2.0}, 4.0, 8.0\}$ \\
			RLAT & $\eps \in \{0.05, \textbf{0.08}, 0.1, 0.15, 0.2, 0.25\}$ \\
			\midrule
			RLAT + DeepAugment & $\eps \in \{0.005, \textbf{0.02}, 0.05\}$ \\
			RLAT + AugMix & $\eps \in \{ 0.01, \textbf{0.02}, 0.05, 0.1, 0.15, 0.2\}$ \\
			RLAT + AugMix + JSD & $\eps \in \{ \textbf{0.01}, 0.02, 0.05, 0.1, 0.15, 0.2\}$ \\
		\end{tabular}
		\caption{Grid searches performed for CIFAR-10 experiments. For each grid, we boldface the hyperparameter that leads to the model with the best common corruption accuracy.}
		\label{tab:app_hyperparams_cifar10}
	\end{table*}
	\begin{table*}[t]
		\centering
		\footnotesize
		\begin{tabular}{@{}ll@{}}
			\textbf{Method} & \textbf{Grid values} \\
			\midrule
			100\% Gaussian augmentation & $\sigma \in \{ \textbf{0.001}, 0.005, 0.01, 0.02, 0.05, 0.1, 0.2, 0.4, 0.5, 0.6\}$ \\
			50\% Gaussian augmentation & $\sigma \in \{ 0.01, 0.02, 0.05, 0.1, 0.2, \textbf{0.5}, 0.8 \}$ \\
			$\l_\infty$ adversarial training & $\eps \in \{ 0.01, \textbf{0.05}, 0.1, 0.5, 1.0, 2.0, 3.0, 4.0 \} / 255$ \\
			$\l_2$ adversarial training & $\eps \in \{ 0.02, \textbf{0.05}, 0.1, 0.2\}$ \\
			RLAT & $\eps \in \{ 0.01, \textbf{0.02}, 0.05, 0.1, 0.2\}$\\
			\midrule
			RLAT + AugMix & $\eps \in \{ 0.01, 0.02, \textbf{0.05}, 0.1, 0.2\}$ \\
			RLAT + AugMix + JSD & $\eps \in \{ 0.01, 0.02, \textbf{0.05}, 0.1, 0.2\}$ \\
			RLAT + SIN & $\eps \in \{0.005, \textbf{0.01}, 0.02, 0.05\}$ \\
			RLAT + ANT\textsuperscript{3x3} & $\eps \in \{ 0.005, 0.01, \textbf{0.02}, 0.05\}$ \\
		\end{tabular}
		\caption{Grid searches performed for ImageNet-100 experiments. For each grid, we boldface the hyperparameter that leads to the model with the best common corruption accuracy.}
		\label{tab:app_hyperparams_imagenet}
	\end{table*}
	
	\myparagraph{Evaluation details.}
	There are 15 corruptions and 5 severity levels in CIFAR-10-C and ImageNet-C. Thus there are multiple ways of reporting the performance of a model on these two datasets. For a model $f$, let $E_{s,c}^{f}$ denote the top-1 error rate on the corruption $c$ with severity level $s$ averaged over the whole test set, then three popular metrics are often reported:
	\begin{itemize}
		\item \textit{Average accuracy}: the accuracy is averaged over all severity levels and corruptions: 
		\[\text{Accuracy}_f = 1 - \frac{1}{15 \cdot 5}\sum_{c=1}^{15} \sum_{s=1}^{5} E_{s,c}^{f}.\]
		\item \textit{Mean corruption error} (mCE, proposed in \cite{hendrycks2019benchmarking}): the error rate on each corruption is normalized by the error rate, $E_{s,c}^{\text{AlexNet}}$, of the standard deep learning model, AlexNet \citep{krizhevsky2012imagenet}:
		\[\text{mCE}_f = \frac{1}{15}\sum_{c=1}^{15} \frac{\sum_{s=1}^{5} E_{s,c}^{f}}{\sum_{s=1}^{5} E_{s,c}^{\text{AlexNet}}}.\] 
		The motivation is to make the error rates on different corruptions more comparable. Indeed they do not all have the same inherent level of difficulty.
		\item \textit{Relative mean corruption error} (relative mCE, proposed in \cite{hendrycks2019benchmarking}): instead of measuring the error rate, one can also consider the degradation of the error rate compared to the standard error rate $E_{\text{standard}}^{f}$ of the model $f$ taken relative to the degradation of the error rate $E_{\text{standard}}^{\text{AlexNet}}$ of AlexNet:
		\[\text{Relative mCE}_f = \frac{1}{15}\sum_{c=1}^{15} \frac{\sum_{s=1}^{5} E_{s,c}^{f} - E_{\text{standard}}^{f}}{\sum_{s=1}^{5} E_{s,c}^{\text{AlexNet}} - E_{\text{standard}}^{\text{AlexNet}}}.\] 
		However, this metric has to be carefully interpreted since it does not take absolute accuracy into account, e.g., a \textit{constant} model achieves the perfect score of 0 for this metric.
	\end{itemize}
	Since there is no standard AlexNet model on CIFAR-10, the mean corruption error and the relative mean corruption error are not well defined on this dataset. Therefore
	we focus on reporting average accuracy on both CIFAR-10-C and ImageNet-100-C to be consistent throughout the paper.
	
	For the LPIPS robustness evaluation shown in Fig.~\ref{fig:lpips_robustness}, we use the following settings: Fast Lagrangian Attack from \cite{laidlaw2021perceptual} using their suggested hyperparameters and AlexNet as the network to compute the LPIPS distance. 
	For the $\l_2$ robustness evaluation, we use the APGD-CE attack \citep{croce2020reliable} with 100 iterations and 5 random restarts.

	\myparagraph{Training details.}
	In all our experiments, we use SGD with momentum to train a PreAct ResNet-18 network (both on CIFAR-10 and ImageNet-100). The momentum coefficient is set to the value $0.9$. The learning rate is initially set to the value $0.1$ and then is decayed  by a factor $10$ according to a predefined schedule.
	We train for 150 epochs on CIFAR-10 and 100 epochs on ImageNet-100 and always report the results of the \textit{last} model, i.e. we do not perform any early stopping.
	We specify all the main training hyperparameters in Table~\ref{tab:app_hyperparams_main}.

	A recent work of \cite{mintun2021interaction} suggests that the most effective data augmentations are those which are perceptually similar to the target corruptions from CIFAR-10-C and ImageNet-C.
	Along the same lines, \cite{rusak2020simple} mention that for AugMix, there is a visual similarity, e.g., between the posterize operation and the JPEG corruption.
	Thus, to prevent training on augmentations which resemble the ones from CIFAR-10-C and ImageNet-C, we use only random horizontal flip and random crops unless mentioned otherwise.
	Moreover, when we train Fast PAT on CIFAR-10, we make sure to remove the overlapping augmentations used in the robustness library \citep{engstrom2019robustness} such as random brightness and contrast change. 
	For the experiments whose results are reported in Table~\ref{tab:main_cifar10_imagenet100}, we use additional augmentations like AugMix, SIN, ANT\textsuperscript{3x3} whenever it is explicitly mentioned. 

	For every method that we reported, we performed a grid search over the main hyperparameters such as the standard deviation $\sigma$ for Gaussian data augmentation or the perturbation radius $\eps$ for adversarial training. We report all the used grids in Table~\ref{tab:app_hyperparams_cifar10} and Table~\ref{tab:app_hyperparams_imagenet}. We note that the grids are not of the same size for all methods since in case an initial grid of values came out to be suboptimal, we expanded it further until the optimal value (according to common corruption accuracy) was attained not at the boundary of the grid. The only exception is for 100\% Gaussian augmentation on ImageNet-100 in Table~\ref{tab:app_hyperparams_imagenet} where the best performance is attained for the smallest $\sigma=0.001$ out of the final grid. We found out that even such a small $\sigma$ still harms the overall performance due to $\sigma$-overfitting and elaborate further on this phenomenon on ImageNet-100 in Sec.~\ref{sec:app_sigma_overfitting_details}.

	\myparagraph{Data augmentation experiments.}
	For the experiments with DeepAugment \citep{hendrycks2020many}, we generate distorted images once before training using the CAE model from their public repository.
	For the experiments that involve training on Stylized ImageNet, we use in each batch 28 stylized images and 100 standard ImageNet images following \cite{rusak2020simple}.
	ANT \citep{rusak2020simple} is the only exception where instead of training from a random initialization, we follow the scheme of the authors and fine-tune a standardly pretrained ImageNet-100 model.
	Additionally, we note that the noise generator of \cite{rusak2020simple} uses skip connections with Gaussian noise and experience replay of previous noise generators. 
	This means that there is a certain Gaussian noise component in the final noise which implies that it partially overlaps with the common corruptions from CIFAR-10-C and ImageNet-100-C.

	\myparagraph{Runtime of RLAT.}
	\begin{table}[t]
		\centering
		\footnotesize
		\begin{tabular}{@{}lrr@{}}
			& \multicolumn{2}{c}{\textbf{Dataset}} \\
			\cmidrule(lr){2-3}
			\textbf{Training} & CIFAR-10 & ImageNet-100 \\
			\midrule
			Standard & 0.8h & 3.9h \\
			$\l_2$/$\l_\infty$ adversarial & 1.3h & 5.8h\\
			RLAT & 1.8h & 6.2h\\
			Fast PAT & 9.4h & \textsuperscript{*}120h\\
		\end{tabular}
		\caption{Wall-clock time in hours for PreAct ResNet-18 trained with different methods on CIFAR-10 and ImageNet-100 using one Nvidia V100 GPU. \textsuperscript{*} denotes the time reported by \cite{laidlaw2021perceptual} for a larger model (ResNet-50) using different hardware (4 Nvidia RTX 2080 Ti GPUs). 
		}
		\label{tab:app_execution_time}
	\end{table}
	We report a full runtime comparison between standard training, $\l_2$ / $\l_\infty$ adversarial training, RLAT, and Fast PAT in Table~\ref{tab:app_execution_time}. The main observation is that RLAT is significantly faster than Fast PAT and leads only to a slight overhead compared to $\l_2$ / $\l_\infty$ adversarial training.

	\myparagraph{Licenses for the used and released assets.}
	We release all our models under the MIT license.
	Throughout the paper, we used ImageNet \citep{russakovsky2015imagenet}, CIFAR-10 \citep{krizhevsky2009learning}, ImageNet-C \citep{hendrycks2019benchmarking}, and CIFAR-10-C \citep{hendrycks2019benchmarking} datasets and the Fast PAT \citep{laidlaw2021perceptual} model trained on ImageNet. Their licenses can be found in their repositories or webpages. Importantly, all their licenses are compatible for the purposes of academic research.

	\section{Additional details and results on calibration}
	\label{sec:app_calibration}
	In this section, we discuss the details on how we compute the expected calibration error and perform temperature rescaling to improve calibration. We also show calibration results for $\l_2$ adversarially trained models.
	
	\myparagraph{Calibration details.}
	To compute the expected calibration error (ECE) we follow the code of \cite{guo2017calibration} with their default settings using 15 equally-sized bins to compute the calibration error. 
	However, we change the implementation of the temperature rescaling. Since optimization of ECE over the softmax temperature is a simple one-dimensional optimization problem, it can be solved efficiently using a grid search. Moreover, we can optimize directly the metric of interest, i.e. ECE, instead of the cross-entropy loss as in \cite{guo2017calibration} who relied on a differentiable loss since they used gradient descent to optimize the temperature.
	As the grid, we use the interval $t \in [0.001, 1.0]$ with a grid step $0.001$ and we test both $t$ and $1/t$ temperatures. Moreover, we make sure that for all methods the optimal $t$ is located not at the boundary of the grid.
	We optimize the temperature only on the \textit{in-distribution} samples from the test sets of CIFAR-10 and ImageNet-100 to make sure that the out-distribution samples from CIFAR-10-C and ImageNet-100-C stay unseen.
	
	\begin{wrapfigure}{r}{0.43\textwidth}
		\vspace{-9mm}
		\begin{center}
			\includegraphics[width=0.435\columnwidth]{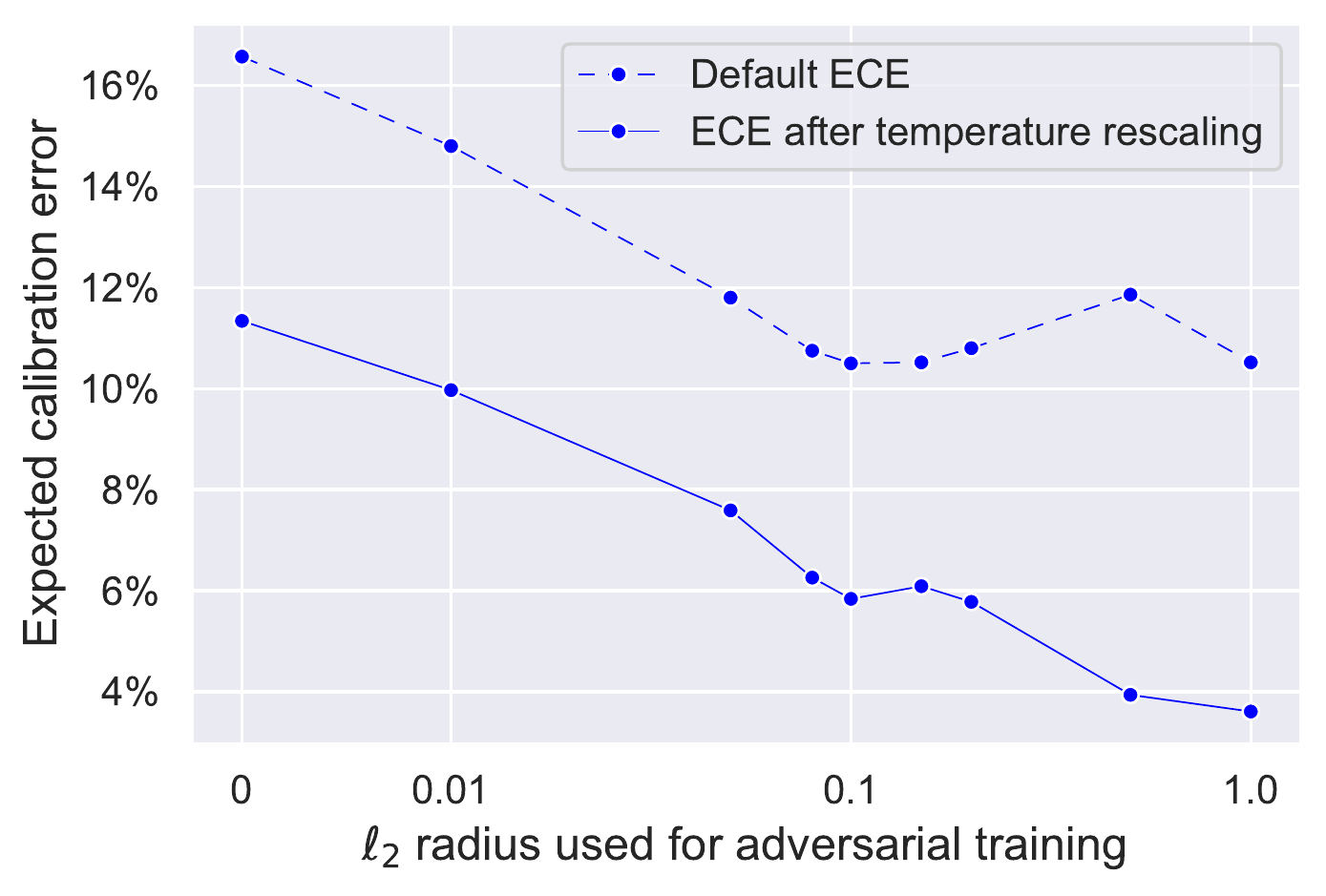}
			\vspace{-6.5mm}
			\caption{Expected calibration error on CIFAR-10-C for $\l_2$ adversarially trained models.}
			\label{fig:calibration_l2}
		\end{center}
		\vspace{-7mm}
	\end{wrapfigure}
	\myparagraph{Calibration of $\l_2$ adversarially trained models.}
	In Fig.~\ref{fig:calibration_l2}, we additionally present the expected calibration error for models adversarially trained with different $\l_2$ perturbation radii. 
	We observe a decreasing trend over the perturbation radius similarly to the $\l_\infty$-trained models shown in Fig.~\ref{fig:calibration_linf}.
	We see that the most accurate $\l_2$ model on corruptions is significantly better calibrated: $10.5\%$ ECE vs $16.6\%$ ECE of the standard model.
	Moreover, the same amount of improvement can be observed even after temperature rescaling: $5.8\%$ ECE vs $11.3\%$ ECE of the standard model.
	Thus, we conclude that both $\l_\infty$ and $\l_2$ adversarial training substantially improve calibration both before and after temperature rescaling. 
	Moreover, $\l_2$ adversarial training leads to better calibration than $\l_\infty$: $10.8\%$ vs $10.5\%$ ECE by default and $6.7\%$ vs $5.8\%$ ECE after temperature rescaling.

	\section{Ablation study for AdvProp and adversarial training}
	\label{sec:app_ablation_advprop_25_50_75_100}
	In this section, we provide experimental details for the AdvProp baseline and compare it with the standard $\l_\infty$ adversarial training.
	
	AdvProp \citep{xie2020adversarial} is a method based on adversarial training where the objective consists of a mixture of clean and adversarial examples for which separate BatchNorm layers are used, and only the clean BatchNorm layers are used at test time. This method was shown to improve the accuracy on clean images compared to standard adversarial training and to help to generalize under distribution shifts such as common corruptions, so we consider it here in more detail.
	As shown in Table~\ref{tab:main_cifar10_imagenet100}, AdvProp achieves 94.7\% standard accuracy and $82.9\%$ common corruption accuracy. Thus AdvProp performs comparably to 100\% $\l_\infty$ adversarial training ($82.9\%$ vs $82.7\%$ accuracy on CIFAR-10-C), however, AdvProp still performs worse than 100\% $\l_2$ adversarial training ($83.4\%$) and 100\% RLAT ($84.1\%$).
	\begin{figure}[h]
		\centering
		\includegraphics[width=0.55\columnwidth]{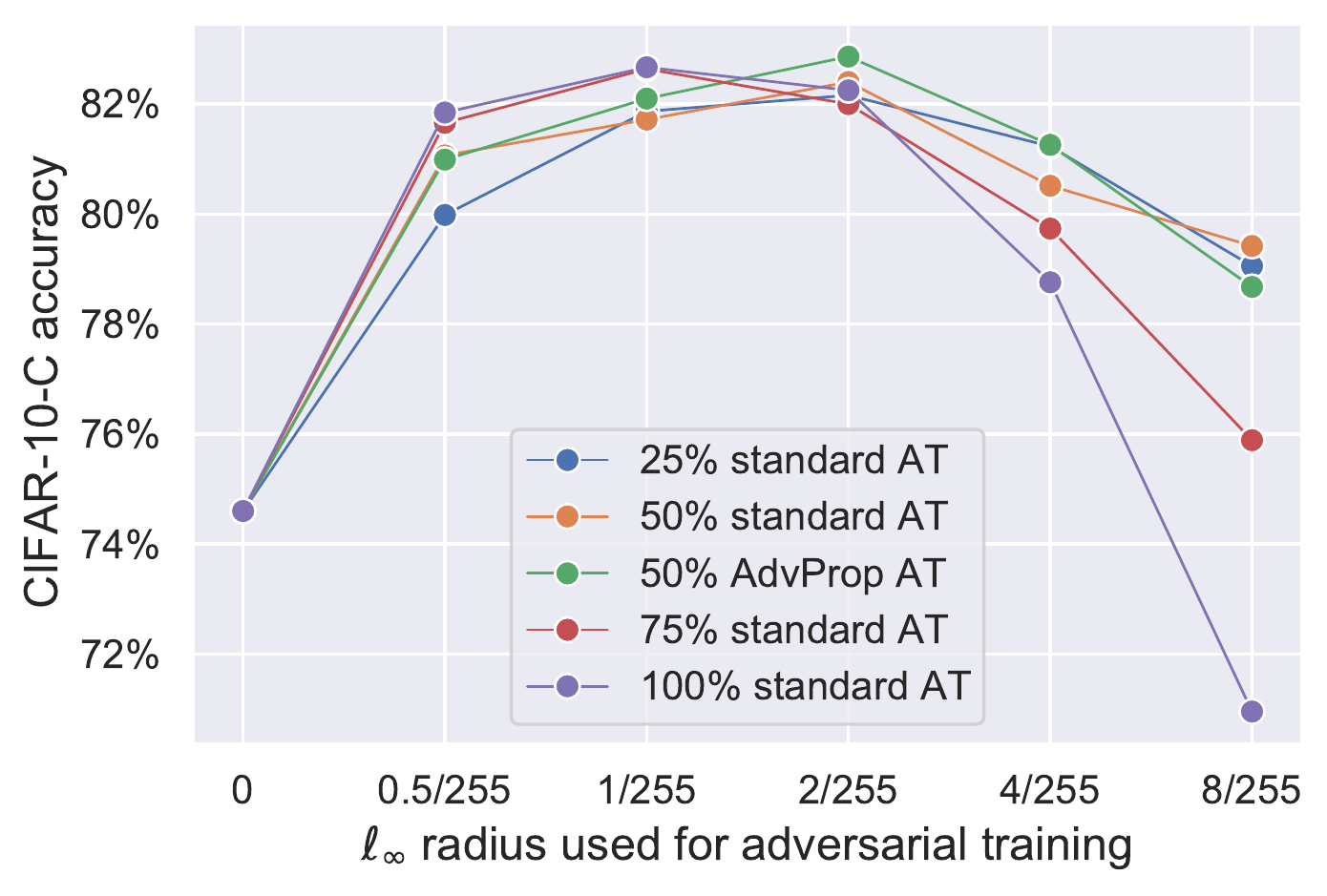}
		\vspace{-3mm}
		\caption{
			Accuracy on common corruptions from CIFAR-10-C for ResNet-18 models adversarially trained using different $\l_\infty$ radii and different proportions of adversarial and clean examples together with the AdvProp scheme.} 
		\label{fig:app_linf_eps_c10c_acc_ablation}
	\end{figure}

	In Fig.~\ref{fig:app_linf_eps_c10c_acc_ablation}, we show the results of an ablation study for PreAct ResNet-18 models adversarially trained using different $\l_\infty$ radii and different proportions of adversarial and clean examples ($25\%$, $50\%$, $75\%$, $100\%$) together with the AdvProp scheme. 
	We can see that AdvProp outperforms 100\% standard AT but only by a small margin (+0.2\%). Moreover, $100\%$ standard AT performs comparably to $75\%$ standard AT and better than $50\%$ and $25\%$ standard AT. Thus, we observe no benefit in mixing clean and adversarial samples for standard AT unlike for Gaussian data augmentation.
	We emphasize here that the advantage of the standard $\l_p$ adversarial training is that it is a conceptually simpler method as it does not require using separate BatchNorms during training, and balancing clean and adversarial samples. 
	Besides, AdvProp requires up to 50\% more training time if the same number of adversarial examples is used as for 100\% standard AT.

	\section{Details on $\sigma$-overfitting}
	\label{sec:app_sigma_overfitting_details}
	In this section, we provide experiments related to the $\sigma$-overfitting phenomenon on ImageNet-100 and provide a further discussion on it.
	\begin{figure*}[t]
		\begin{center}
			\begin{subfigure}[t]{.44\textwidth}
				\centering 
				\includegraphics[width=0.89\columnwidth]{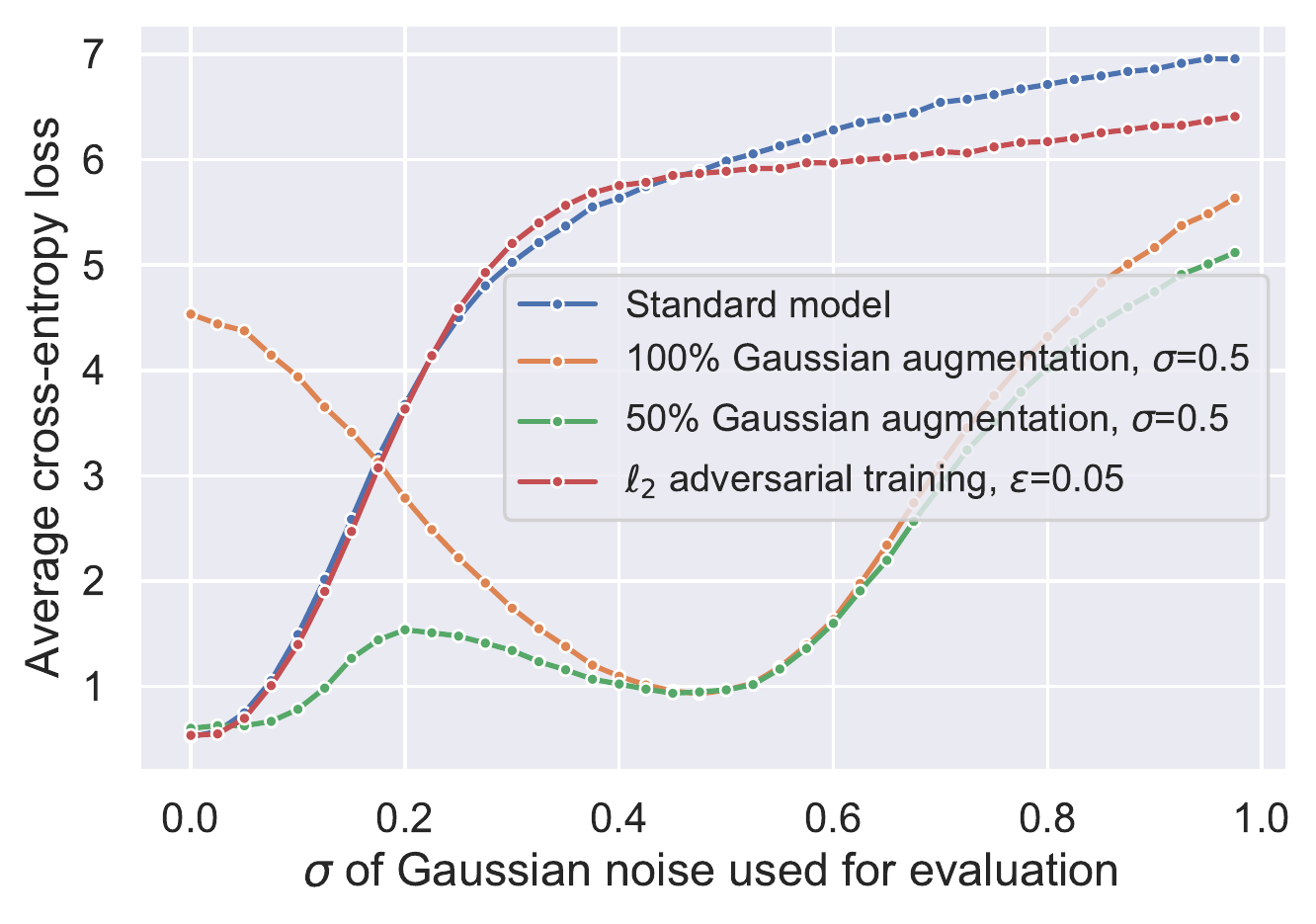}
				\caption{A larger range of the standard deviation $\sigma \in [0, 1]$}
			\end{subfigure}
			\hspace{4mm}
			\begin{subfigure}[t]{.47\textwidth}
				\centering 
				\includegraphics[width=0.85\columnwidth]{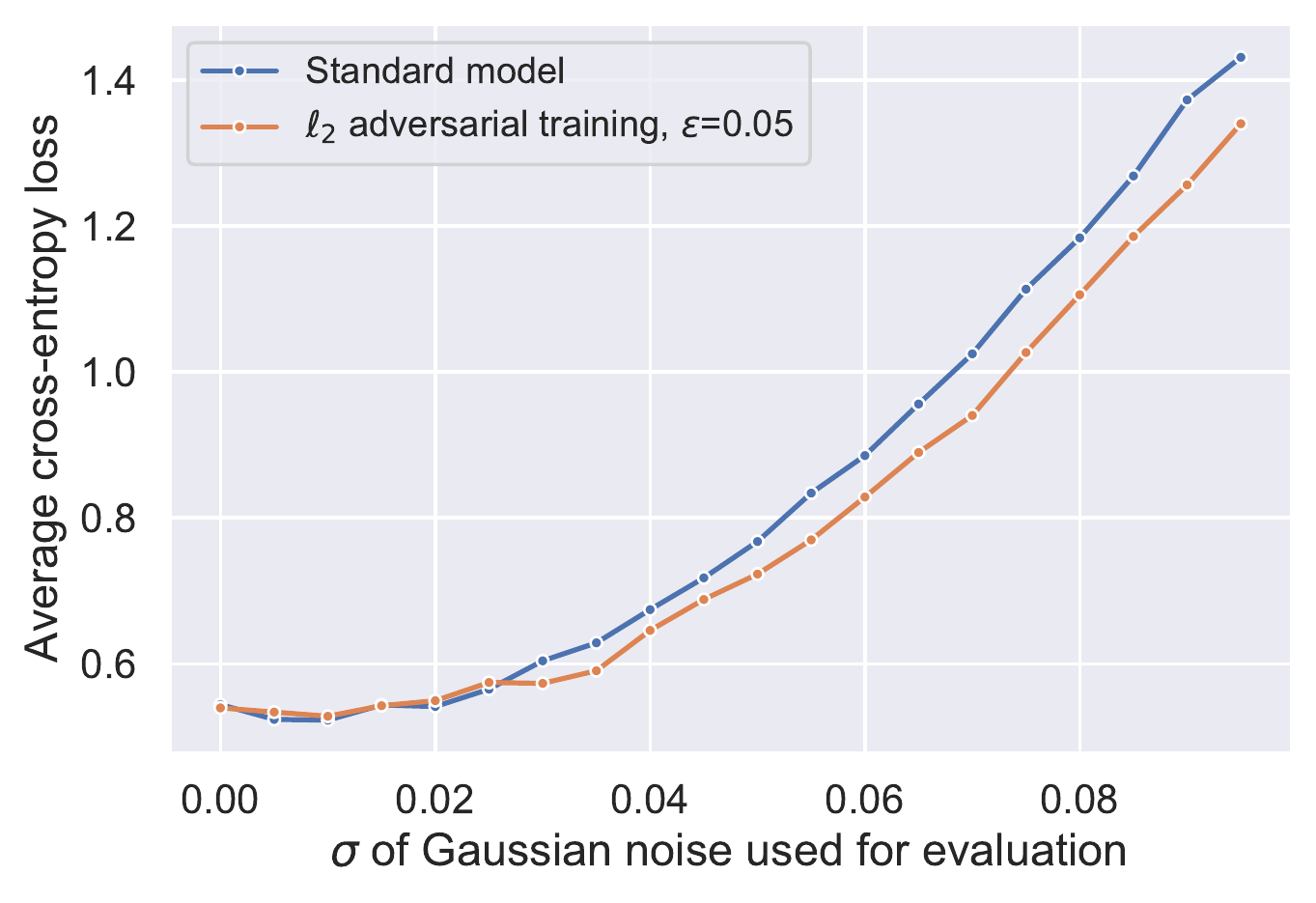}
				\caption{A smaller range of the standard deviation $\sigma \in [0, 0.1]$}
			\end{subfigure}
			\caption{The cross-entropy loss under Gaussian noise of different training methods for ImageNet-100. We see that 100\% Gaussian augmentation \textit{severely} overfits to the noise magnitude used for training while the 50\% Gaussian augmentation scheme mitigates this problem, but not completely, since it has a noticeable local maximum $\sigma=0.2$, and $\l_2$ adversarial training does not suffer from this problem.}
			\label{fig:sigma_overfitting_imagenet}
		\end{center}
	\end{figure*}

	\myparagraph{$\sigma$-overfitting on ImageNet.}
	On ImageNet-100-C, the gap between $50\%$ Gaussian augmentation and $100\%$ Gaussian augmentation is even larger than on CIFAR-10-C (see Table.~\ref{tab:main_cifar10_imagenet100}). 
	To study the $\sigma$-overfitting phenomenon in more detail, we repeat the experiment behind Fig.~\ref{fig:sigma_overfitting} on ImageNet-100 and show the results in Fig.~\ref{fig:sigma_overfitting_imagenet}.
	
	We first focus on large Gaussian perturbations (up to $\sigma=1$ for image pixels in $[0, 1]$) in Fig.~\ref{fig:sigma_overfitting_imagenet}~(a). 
	We observe that 100\% Gaussian augmentation \textit{severely} overfits to the noise magnitude used for training while the 50\% Gaussian augmentation scheme mitigates this problem.
	However, 50\% Gaussian augmentation does not completely solve the $\sigma$-overfitting problem since we observe that the loss still has a local maximum around $\sigma=0.2$ (with a $\approx1.6\times$ increase compared to the loss at $\sigma=0.45$).
	Therefore, 50\% Gaussian augmentation may be a suboptimal method against $\sigma$-overfitting. 
	We believe that future research is needed to better understand improved mitigation strategies.

	We additionally plot the performance of standard and $\l_2$ adversarially trained models for a smaller range of Gaussian perturbations with $\sigma \in [0, 0.1]$ in Fig.~\ref{fig:sigma_overfitting_imagenet}~(b). We can observe that there is no $\sigma$-overfitting trend for standard and $\l_2$ adversarially trained models. Moreover, the latter has a slightly smaller loss for small values of $\sigma$.

	\myparagraph{\citet{ford2019adversarial} in the context of $\sigma$-overfitting.}
	As a Gaussian data augmentation baseline, \cite{ford2019adversarial} use a scheme that is actually different from both 50\% and 100\% Gaussian augmentation schemes. 
	They perform Gaussian augmentation on \textit{each} sample, however they sample the standard deviation uniformly at random from the range $[0, \sigma]$\footnote{We confirmed this implementation detail via private communication with the authors. Note that this also corresponds to what \citep{rusak2020simple} report in Appendix~H.}. This strategy can be seen as an interpolation between the 50\% and 100\% Gaussian augmentation schemes. Moreover, another difference in the results from \cite{ford2019adversarial} compared to our paper is that they computed corruptions for ImageNet-C in-memory which leads to some discrepancy compared to static images (e.g., see Fig.~5 in \cite{ford2019adversarial} for an illustration). 
	
	\myparagraph{$\sigma$-overfitting and adversarial training.}
	As we can see from Fig.~\ref{fig:sigma_overfitting} and Fig.~\ref{fig:sigma_overfitting_imagenet}, adversarial training does not suffer from the $\sigma$-overfitting problem which may explain why applying 50\% adversarial training (as in Sec.~\ref{sec:app_ablation_advprop_25_50_75_100}) does not lead to better results compared to 100\% adversarial training. 
	We believe that the difference between Gaussian augmentation and adversarial training can occur due to a much larger norm of the perturbation used for data augmentation.
	On CIFAR-10-C, the optimal $\l_2$ radius for adversarial training is $0.1$ while the norm of Gaussian vectors is approximately $\sigma_{100\%} \cdot \sqrt{d} = 0.05 \cdot \sqrt{32 \cdot 32 \cdot 3} \approx 2.77$ for 100\% Gaussian augmentation and $\sigma_{50\%} \cdot \sqrt{d} = 0.1 \cdot \sqrt{32 \cdot 32 \cdot 3} \approx 5.54$ for 50\% Gaussian augmentation.
	It is therefore likely that due to a much larger norm of the perturbation, the model trained with 100\% Gaussian noise fails to generalize to perturbations of smaller norms, while this failure can be alleviated to some extent by using the 50\% scheme. 
	On a related note, some variations of adversarial training are known to suffer from a phenomenon called \textit{catastrophic overfitting} \citep{tramer2018ensemble, wong2020fast}. Catastrophic overfitting shares some similarities with $\sigma$-overfitting, in the sense that the model overfits to a particular perturbation type used during training.  Clarifying the relation of catastrophic overfitting and $\sigma$-overfitting is an interesting direction for future work.
	

	\section{Suitability of LPIPS for common corruptions.}
	\label{sec:app_suitability_of_lpips}
	We first show that the LPIPS distance is more suitable to common image corruptions than the $\l_2$ distance. We note that previous works \citep{zhang2018unreasonable, laidlaw2021perceptual} have discussed its advantages over other commonly used distances, however not in the context of corruptions from CIFAR-10-C. Another difference to \cite{zhang2018unreasonable} is that here we compute the LPIPS distance using \textit{full images} instead of smaller image patches as we are eventually interested in performing adversarial training using full inputs.
	
	We compute the average $\l_2$ and LPIPS distances based on a standardly trained VGG network using the code from \cite{zhang2018unreasonable} for different corruptions from CIFAR-10-C. The results are shown in Fig.~\ref{fig:heatmap_distances_corruptions} (see also Table~\ref{tab:cc_l2lpipsdist} in the Appendix), where we observe that for certain corruptions, LPIPS clearly demonstrates a more preferable behavior than the $\l_2$ distance.
	For example, the $\l_2$ distances of elastic transformations are monotonically \textit{decreasing} over the corruption severity which is the opposite of what we would expect from a suitable distance between images. At the same time, the LPIPS distance is first slightly decreasing and then increasing which is a better behavior compared to $\l_2$.
	Similarly, the $\l_2$ distances for JPEG corruption appear to be roughly constant while they are noticeably increasing for LPIPS.
	The LPIPS distances for the frost corruption start from a low value and then monotonically increase whereas the $\l_2$ distances start from a very high value and then fail to be monotonic. 
	At the same time, LPIPS behavior is more unexpected on noise corruption where it shows a very fast increase for impulse, Gaussian, and shot noises.

	Therefore, the LPIPS distance appears to better capture the severity of these corruptions. To investigate further this qualitative assessment, we compute the correlation between the $l_2$ and the LPIPS distances and the error rates of a standardly trained model (shown in Fig.~\ref{fig:heatmap_err_clean_model_corruptions}) in Table~\ref{tab:correlations_distances_errors}. More precisely, each corruption has five severity levels for which both the average distance value and the average error rate can be calculated. Therefore, for a given corruption, we compute the correlation between the vector composed of the distances corresponding to each severity level and the error-rate vector defined similarly. We take then the average correlation over each corruption type.
	We observe that the LPIPS distance is \textit{more correlated} with the error rates for all corruption types except the noise one. The main difference comes from the digital corruptions where LPIPS leverages the monotonic behavior of the frost and elastic transform corruptions.
	Therefore, we conclude that LPIPS is quantifiably more suitable to capture the distance between common image corruptions.
	\begin{table}[t]
		\centering
		\small
		\begin{tabular}{@{}lccccc@{}}
			& \multicolumn{5}{c}{\textbf{Corruption type}} \\
			\cmidrule(lr){2-6}
			\textbf{Metric}     & All & Noise & Blur & Weather & Digital  \\
			\midrule
			$\l_2$ & 0.807 & \textbf{0.998} & 0.986 & 0.835 & 0.561 \\
			LPIPS  & \textbf{0.963} & 0.993 & \textbf{0.994} & \textbf{0.968} & \textbf{0.921} \\
		\end{tabular}
		\caption{Correlation between distances and error rates of a standard model taken across different corruption severities and averaged over multiple corruptions. LPIPS distance is better correlated with the error rates on different corruption severity levels.}
		\label{tab:correlations_distances_errors}
	\end{table}

	\myparagraph{Average perturbation distance for different corruptions.}
	\begin{table*}[t]
		\centering
		\footnotesize
		\begin{tabular}{@{} l rrrrr rrrrr @{}}
			& \multicolumn{5}{c}{\textbf{$\l_2$ distance at different}} &   \multicolumn{5}{c}{\textbf{LPIPS distance at different}} \\
			& \multicolumn{5}{c}{\textbf{severity levels}} &   \multicolumn{5}{c}{\textbf{severity levels}} \\
			\cmidrule(lr){2-6} \cmidrule(lr){7-11}
			\textbf{Corruption} &  1   &      2 &      3&      4 &     5  &  1   &      2 &      3&      4 &     5 \\
			\midrule
			Shot noise        & 1.67 &  2.35 &  3.68 &  4.22 &  5.13 &  0.089 &  0.143 &  0.245 &  0.284 &  0.341\\
			Motion blur       & 2.40 &  3.51 &  4.33 &  \cellcolor{red!17} 4.32 &  4.97 &  0.059 &  0.114 &  0.166 &  0.166 &  0.211  \\
			Snow              & 2.92 &  5.83 &  6.60 &  9.10 & 12.17 &  0.068 &  0.155 &  0.160 &  0.188 &  0.227 \\
			Pixelate          & 1.25 &  1.77 &  1.97 &  2.48 &  3.04 &  0.130 &  \cellcolor{red!17} 0.060 &  0.074 &  0.138 &  0.202  \\
			Gaussian noise    & 2.19 &  3.26 &  4.31 &  4.83 &  5.34 &  0.027 &  0.217 &  0.294 &  0.328 &  0.359\\
			Defocus blur      & 0.43 &  1.06 &  1.61 &  2.09 &  2.98 &  0.003 &  0.019 &  0.048 &  0.085 &  0.153 \\
			Brightness        & 2.26 &  4.59 &  6.85 &  9.01 & 12.95 &  0.006 &  0.022 &  0.045 &  0.071 &  0.132  \\
			Fog               & 2.81 &  5.54 &  7.18 &  8.45 & 10.20 &  0.022 &  0.086 &  0.147 &  0.215 &  0.332\\
			Zoom blur         & 3.04 &  3.56 &  4.20 &  4.83 &  5.40 &  0.079 &  0.094 &  0.124 &  0.153 &  0.189\\
			Frost             & 6.86 & 10.14 &  11.39 & \cellcolor{red!17} 10.45 & \cellcolor{red!17} 10.26 &  0.077 &  0.141 &  0.208 &  0.214 &  0.266 \\
			Glass blur        & 4.70 &  \cellcolor{red!17} 4.64 &  \cellcolor{red!17} 4.27 &  6.73 & \cellcolor{red!17} 6.28 &  0.253 &  \cellcolor{red!17} 0.248 &  \cellcolor{red!17} 0.235 &  0.333 & \cellcolor{red!17} 0.321 \\
			Impulse noise     & 3.08 &  4.37 &  5.35 &  6.92 &  8.20 &  0.136 &  0.223 &  0.289 &  0.387 &  0.452 \\
			Contrast          & 2.80 &  5.60 &  6.71 &  7.83 &  9.51 &  0.020 &  0.092 &  0.143 &  0.216 &  0.386 \\
			JPEG compression  & 1.59 &  1.95 &  2.07 &  2.20 &  2.38 &  0.073 &  0.108 &  0.121 &  0.134 &  0.153  \\
			Elastic transform & 7.37 & \cellcolor{red!17} 6.79 & \cellcolor{red!17} 6.11 & \cellcolor{red!17} 5.67 & \cellcolor{red!17} 4.76 &  0.168 &  \cellcolor{red!17} 0.152 &  \cellcolor{red!17} 0.151 &  0.175 &  0.198  \\
		\end{tabular}
		\caption{Average $\l_2$ and LPIPS distance for different corruptions and severity levels from CIFAR-10-C. Note that the distances do not always monotonically increase with the corruption level. We mark such cases in {\color{red}\textbf{red}} and observe that they occur less often for LPIPS than for the $\l_2$ distance.}
		\label{tab:cc_l2lpipsdist}
	\end{table*}
	In Table \ref{tab:cc_l2lpipsdist} we report the average distances between clean and corrupted images for the LPIPS and $l_2$ norms which is the same data as in Fig.~\ref{fig:heatmap_distances_corruptions}. We report exact numbers to further illustrate the point that adversarial training with worst-case perturbations in a \textit{small} $\l_2$ ball (such as $\eps=0.1$) leads to robustness against corruptions of a much larger magnitude. We can observe from Table \ref{tab:cc_l2lpipsdist} that for some corruptions, the perturbation norm does not grow monotonically (particularly, glass blur and elastic transform) which we highlight in {\color{red}\textbf{red}}. We observe that for the LPIPS distance such behavior occurs less often. Another observation is that the $l_2$ distance itself does not always accurately reflects the strength of the performance degradation. For example, fog and brightness have similar magnitude (and the largest among the other corruptions), but very different behavior in terms of accuracy: degradation under fog is much higher than under brightness.

	\section{Details on the relaxed LPIPS adversarial training}
	\label{sec:app_lpips_relaxation}
	
	In this section, we first discuss in more detail the approach of \cite{laidlaw2021perceptual}, then present complete derivations for RLAT where LPIPS is defined using multiple layers, discuss implementation details of RLAT, and evaluate the $\l_2$ robustness of RLAT and a few other baselines.

	\myparagraph{Details on the perceptual attacks of \cite{laidlaw2021perceptual}.}
	At each step of the Perceptual Projected Gradient Descent (PPGD) proposed in \cite{laidlaw2021perceptual}, both the loss $\l$ and the neural network $f$ used to define $\lpips$ are linearized, and the constrained problem~\eqref{eq:lpips_at_obj} is approximated with a large linear system which is solved approximately with $K$ iterations of the conjugate gradient method. To satisfy the constraint in~\eqref{eq:lpips_at_obj}, the solution $\delta$ is then approximately projected onto the LPIPS-ball, i.e. onto the set $\{\delta : \lpips(x, x+\delta)\leq\eps\}$, for which $n$ iterations of the bisection method are used. 
	For $T$ iterations of PPGD, the algorithm in total requires $T(K + n + 4)$ forward passes and $T(K + n + 3)$ backward passes of the network which makes it significantly more expensive than standard PGD which requires $T$ forward and backward passes.
	
	The Lagrangian Perceptual Attack (LPA) uses the following Lagrangian relaxation of the objective:
	\begin{align*}
	\max_{\delta} \l(x+\delta, y; \theta) -\lambda \max \{ \lpips(x,x+\delta)-\varepsilon,0\},
	\end{align*}
	which is solved by gradient descent for several values of the Lagrange multiplier $\lambda$ (usually $S=5$ in their experiments) and whose solution is then projected back onto the LPIPS-ball.  
	In total, the attack requires $2ST + n + 2$ forward passes and $ST + n + 2$ backward passes of the network which also makes it expensive due to the outer loop over $S$ different values of $\lambda$.
	
	These two attacks are too computationally expensive to be efficiently used during adversarial training. To speed up the method, they additionally propose Fast-LPA where $\lambda$ is not searched over but is instead increased during the training according to a fixed schedule and no projection steps are included. 
	Then Fast-LPA requires $2T + 1$ forward and $T + 1$ backward passes 
	that represent a small overhead compared to PGD but a large one when compared to single-step methods such as FGSM. 
	We note that the possibility of using Fast-LPA with a few iterations is worth investigating in future work, although it appears to be not straightforward because of the Lagrangian formulation and the need to tune the parameter $\lambda$ over iterations of Fast-LPA.

	\myparagraph{Relaxation for multi-layer LPIPS.}
	We derive here the relaxation of LPIPS adversarial training for a general \textit{multi-layer} version of the LPIPS distance. We recall from Eq.~\eqref{eq:lpips_def} that the LPIPS distance can be written as
	\begin{align*}
	\lpips(x,x+\delta)^2 = \sum_{l=1}^L \alpha_l\|\phi_l(x)-\phi_l(x+\delta)\|_2^2.
	\end{align*}
	We use the convention that $\alpha_l = 0$ if a layer $l$ is not in the set of the layers used in the LPIPS distance, i.e. if $l \not\in \mathcal{L}_{LPIPS}$.
	We consider the network $f$ written in its compositional form, i.e., $f(x) = g_L \circ \cdots \circ g_1(x)$. The LPIPS adversarial problem defined in Eq.~\ref{eq:lpips_at_obj} is then equivalent to the problem 
	\begin{align*}
	\max_{\delta} \quad & \l(g_L(\dots g_1(x + \delta) \dots )) \\
	\text{s.t. } \quad & \sum_{l=1}^L \alpha_l \norm{\phi_l(x)-\phi_l(x+\delta)}_2^2 \leq \eps^2.
	\end{align*}
	We introduce the slack variables $\tilde \delta^{(l)}$ for $l=1,\dots,L$, defined as $\tilde \delta^{(l)} = g_l(g_{l-1}(\dots g_1(x + \tilde \delta^{(1)}) \dots) + \tilde \delta^{(l-1)})  g_l( \dots g_1(x) \dots )$ when $ l\in \mathcal{L}_{LPIPS}$ and  $\tilde \delta^{(l)} = 0$ otherwise. The previous problem can be written as:
	\begin{align*}
	\max_{\tilde \delta^{(1)}, \dots, \tilde \delta^{(L)}} \quad & \l(g_L(\dots g_1(x + \tilde \delta^{(1)}) \dots + \tilde\delta^{(L)})) \\ 
	\text{s.t.} \quad & \sum_{l=1}^L \alpha_l \norm{\tilde \delta^{(l)}}_2^2 \leq \eps^2,  \\
	& \tilde \delta^{(l)} = g_l(g_{l-1}(\dots g_1(x + \tilde \delta^{(1)}) \dots) + \tilde \delta^{(l-1)}) - \\
	& \ \ \ \ \ \ \ \ \ \ \ g_l( \dots g_1(x) \dots ) \quad \forall l \in \mathcal{L}_{LPIPS},  \\
	& \tilde \delta^{(l)} = 0 \quad \forall l \not\in \mathcal{L}_{LPIPS}.
	\end{align*}
	When relaxing the equality constraints on the slack variables $\tilde \delta^{(l)}$ for $l \in \mathcal{L}_{LPIPS} $ we obtain the following relaxation 
	\begin{align*}
	\max_{\tilde \delta^{(1)}, \dots, \tilde \delta^{(L)}} \quad & \l(g_L(\dots g_1(x + \tilde \delta^{(1)}) \dots + \tilde\delta^{(L)}))  \\
	\text{s.t.} \quad & \sum_{l=1}^L \alpha_l \norm{\tilde \delta^{(l)}}_2^2 \leq \eps^2, \\
	\quad & \tilde \delta^{(l)} = 0 \quad \forall l \not\in \mathcal{L}_{LPIPS}.  
	\end{align*}
	We further relax the inequality constraint on  $ \sum_{l=1}^L \alpha_l \norm{\tilde \delta^{(l)}}_2^2$ as individual constraints on each $\norm{\tilde \delta^{(l)}}_2$ in the following way
	\begin{align*}
	\max_{\tilde \delta^{(1)}, \dots, \tilde \delta^{(L)}} \quad & \l(g_L(\dots g_1(x + \tilde \delta^{(1)}) \dots + \tilde\delta^{(L)})) \\
	\text{s.t.} \quad & \norm{\tilde \delta^{(l)}}_2 \leq \frac{\eps}{\sqrt{\alpha_l}}  \quad  \forall l \in \mathcal{L}_{LPIPS}, \\
	\quad & \tilde \delta^{(l)} = 0  \quad\quad\quad\quad \ \forall l \not\in \mathcal{L}_{LPIPS}.
	\end{align*}
	Denoting by $ \eps_l = \frac{\eps}{\sqrt{\alpha_l}}$, we finally obtain the RLAT relaxation from Eq.~\ref{eq:lpips_relax_multilayer}
	\begin{align}
	\label{eq:rlatfinal}
	\max_{\tilde \delta^{(1)}, \dots, \tilde \delta^{(L)}} \quad & \l(g_L(\dots g_1(x+\tilde\delta^{(1)}) \dots + \tilde\delta^{(L)}))  \\ 
	\text{s.t.} \quad & \| \tilde \delta^{(l)} \|_2\leq \varepsilon_l  \quad  \forall l \in \mathcal{L}_{LPIPS}, \nonumber \\
	\quad & \tilde \delta^{(l)} = 0  \quad\quad \ \ \forall l \not\in \mathcal{L}_{LPIPS}. \nonumber 
	\end{align}

	We provide the algorithm for a single iteration of weight updates for RLAT in Algorithm~\ref{alg:rlat}. We show the weight update for standard SGD but any other optimizer can be used as well.
	We emphasize that one of the important advantages of RLAT is its computational efficiency since it leads to only $2\times$ overhead since we can successfully use a single-step adversarial training for it (see Fig.~\ref{fig:lpips_robustness} for LPIPS robustness evaluation with an iterative attack). We refer to Table~\ref{tab:app_execution_time} for exact timings and comparison to Fast PAT and other methods.
	
	\begin{algorithm2e}[t]
		\Indm
		\caption{Single iteration of relaxed LPIPS adversarial training (RLAT)}
		\label{alg:rlat}
		\SetKwInOut{Input}{\hspace{4.5mm} input}
		\SetKwInOut{Output}{\hspace{4.5mm} output}
		\Input{network weights $\theta$, batch of training samples $(x_i, y_i)_{i=1}^b$}
		\Output{updated network weights $\theta_{new}$}
		\Indp
		\For{$i$ \textbf{in} $\{1, \dots, b\}$}{  
			$\forall l \in \{1, \dots, L\}$: $\tilde\delta_i^{(l)} := 0$ \\  
			\For{$l$ \textbf{in} $\mathcal{L}_{LPIPS}$}{
				$\nabla_i^{(l)} := \nabla_{\tilde\delta_i^{(l)}} \l(g_L(\dots g_1(x_i+\tilde\delta_i^{(1)}) \dots + \tilde\delta_i^{(L)}), y_i)$ \\
			}
			\For{$l$ \textbf{in} $\mathcal{L}_{LPIPS}$}{  
				$\tilde\delta_i^{(l)} := \eps_l \nabla_i^{(l)} / \norm{\nabla_i^{(l)}}_2$ \\
			}
		}
		$\theta_{new} := \theta - \eta \nabla_\theta \frac{1}{b} \sum_{i=1}^b \l(g_L(\dots g_1(\Pi_{[0, 1]^d} [ x_i+\tilde\delta_i^{(1)} ]) \dots + \tilde\delta_i^{(L)}), y_i)$ \\
		{\rm \bf return} $\theta_{new}$
	\end{algorithm2e}

	\myparagraph{Layer selection.}
	We choose to use the following layers for LPIPS used for RLAT: input, conv1, conv2\_x, conv3\_x, conv4\_x, and conv5\_x. 
	We tried various combinations of layers including all layers in the network, all BatchNorm layers, and all convolution layers, and the best results were obtained when perturbations are added to the outputs of residual blocks and the first convolution layer.
	
	\myparagraph{Magnitude of layerwise perturbations.}
	Similarly to $\{\alpha_l\}_{l=1}^L$ in the definition of the LPIPS distance in Eq.~\eqref{eq:lpips_def}, the constraints $\{\eps_l\}_{l=1}^L$ for different layers in Eq.~\eqref{eq:rlatfinal} also need to be carefully selected so that the perturbation magnitudes on different layers are balanced.
	Our final approach sets the layerwise bounds $\eps_l$ proportionally to the layer's dimensionality and depth: 
	\[\eps_l = \frac{1}{l} \frac{d_l}{d_{in}} \eps,\] 
	where $d_l$ is the dimension of the $l$-th feature maps and $d_{in}$ is the input dimension.
	We choose this scaling since it is simple enough and empirically more effective than other simple scaling methods that we have tried such as the constant or inverse proportional strategies, or even more involved methods such as dynamic adjusting of the scale to the average of the layer's BatchNorm.
	
	\myparagraph{$\l_2$ robustness of RLAT.}
	\begin{figure}[t!]
		\begin{center}
			\includegraphics[width=0.55\columnwidth]{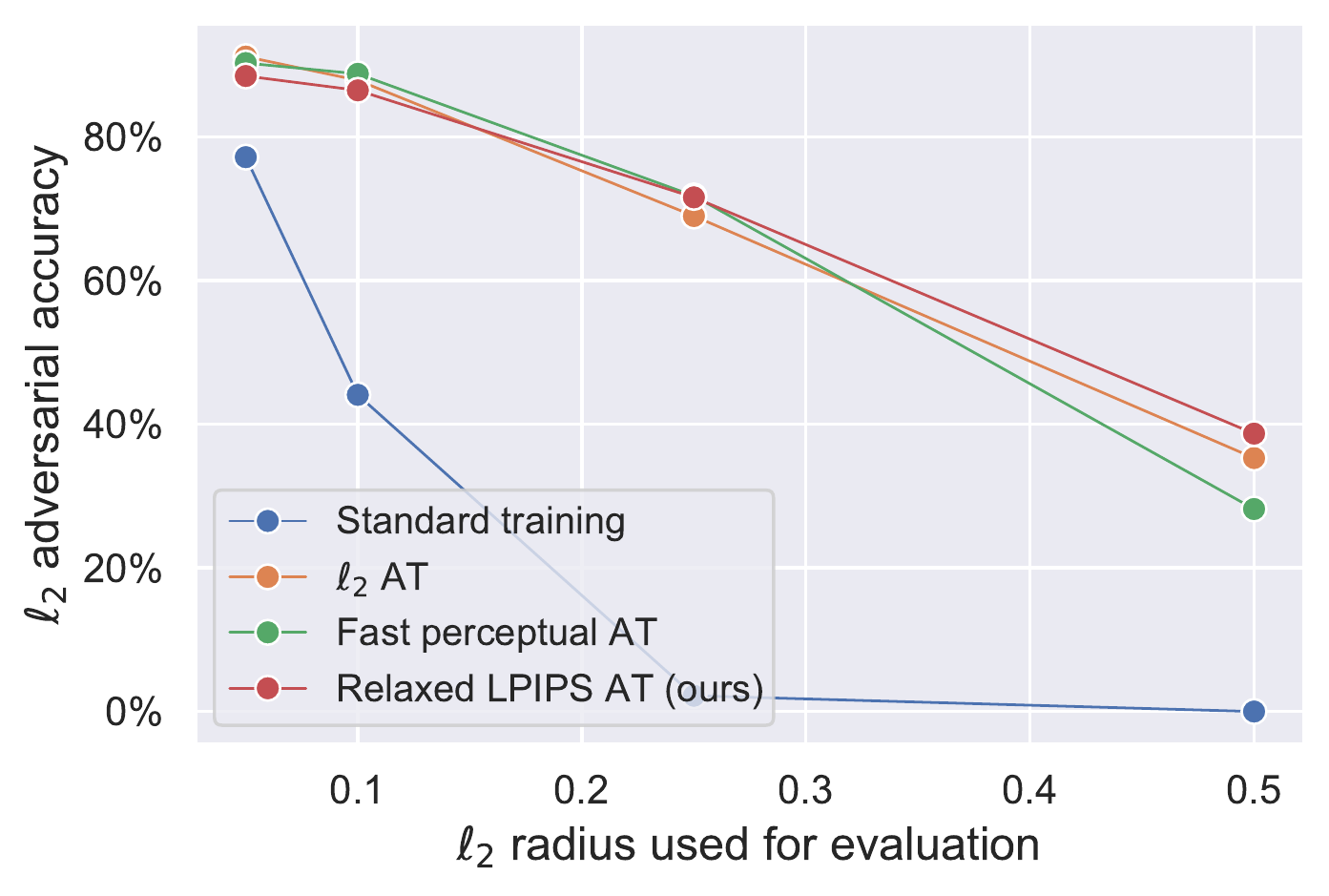}
			\caption{$\l_2$ adversarial robustness of different training schemes on CIFAR-10. All three adversarial training methods improve the $\l_2$ robustness substantially compared to the standard model.}
			\label{fig:l2_robustness}
		\end{center}
	\end{figure}
	To complement the LPIPS robustness evaluation in Fig.~\ref{fig:lpips_robustness}, we also report the $\l_2$ robustness of the same set of models in Fig.~\ref{fig:l2_robustness}: standard, $\l_2$ AT, Fast perceptual AT, and RLAT models with their main hyperparameters selected to perform best on common corruptions. We evaluate $\l_2$ robustness using the APGD-CE attack with 100 iterations and 5 random restarts \citep{croce2020reliable} for different $\l_2$ radii $\eps \in \{0.05, 0.1, 0.25, 0.5\}$. We observe that \textit{all} three adversarial training methods improve the $\l_2$ robustness substantially compared to the standard model.

	\section{Performance under various distribution shifts}
	\label{sec:app_distr_shifts}
	\begin{table*}[t]
		\centering
		\footnotesize
		\begin{tabular}{@{}lrrrr@{}}
			\textbf{Method}                      & \textbf{Standard} & \textbf{IN-100-A} & \textbf{IN-100-R} & \textbf{IN-100-Stylized} \\  
			\midrule
			Standard training           & \textbf{86.6\%} & 5.9\% & {33.2\%} & 16.6\% \\  
			100\% Gaussian augmentation & 86.4\% & 5.8\% & 31.2\% & 17.1\% \\  
			50\%  Gaussian augmentation & 83.8\% & 5.7\% & 32.6\% & \textbf{18.9\%} \\  
			Fast PAT & 71.5\% & 5.4\% & \textbf{34.6}\% & 17.7\% \\
			$\l_\infty$ AT              & 86.5\% & 5.0\% & {33.2\%} & 18.1\% \\  
			$\l_2$ AT                   & 86.3\%  & 5.5\% & {33.2\%} & 17.7\% \\  
			RLAT           & 86.5\% & \textbf{6.3\%} & 33.1\% & 17.7\% \\  
			\midrule
			AugMix            & 86.7\% & \textbf{5.5\%} & 31.5\% & 20.1\% \\  
			AugMix + RLAT   & \textbf{86.8\%} & 5.1\% & \textbf{33.2\%} & \textbf{20.6\%} \\  
			\midrule
			Stylized ImageNet & \textbf{86.6\%} & \textbf{6.5\%} & 35.1\% & \color{gray}62.5\%\color{black} \\  
			Stylized ImageNet + RLAT  & 86.5\% & \textbf{6.5\%} & \textbf{37.0\%} & \color{gray}\textbf{63.4\%}\color{black} \\  
			\midrule
			ANT\textsuperscript{3x3} & \textbf{85.9\%} & \textbf{5.6\%} & \textbf{33.3\%} & \textbf{20.8\%} \\  
			ANT\textsuperscript{3x3} + RLAT & 85.3\% & 5.3\% & 32.8\% & 20.7\%  \\  
		\end{tabular}
		\caption{Accuracy of various methods on different distribution shifts: ImageNet-100-A, ImageNet-100-R, and Stylized ImageNet-100. Gray-colored numbers correspond to models trained and evaluated using Stylized ImageNet.}
		\label{tab:app_imagenet100_domain_shifts}
	\end{table*}
	In this section, we provide additional experiments on distribution shifts that are different from the common corruptions that we studied throughout this paper.
	
	For this, we use three distribution shifts to evaluate our models: ImageNet-A, ImageNet-R, and Stylized ImageNet (SIN). We report the results in Table~\ref{tab:app_imagenet100_domain_shifts} where we compute the accuracy on Stylized ImageNet on all 100 classes, and the accuracy on ImageNet-A and ImageNet-R on all classes that overlap with the classes of ImageNet-100.
	We use the same models for this evaluation as the ones reported in Table~\ref{tab:main_cifar10_imagenet100}, i.e. these models have been selected after a grid search to maximize the performance on ImageNet-100-C.
	We observe that for evaluations on ImageNet-100-A and Stylized ImageNet-100, RLAT moderately improves the accuracy ($+0.4\%$ and $+1.1\%$ respectively) but does not yield improvements on ImageNet-100-R ($-0.1\%$).
	As expected, training on SIN gives very significant improvements for an evaluation on SIN since the same distribution was used during training and testing. 
	We also note that the performance of all methods could be improved if the model selection was performed on the target datasets, and not on ImageNet-100-C.
	Finally, we observe that for some data augmentation methods like AugMix and SIN, using RLAT leads to further improvements, e.g. from 35.1\% to 37.0\% for SIN + RLAT on ImageNet-100-R.
	Overall, we conclude that there is no method that performs best on all distribution shifts, and the obtained improvements are relatively small unless one uses a target distribution shift for training.

	\section{Supplementary figures and tables}
	\label{sec:add_exps_app}
	In this section, we present additional experimental results related to Sections~\ref{sec:at_vs_gauss} and~\ref{sec:main_exps}.

	\myparagraph{Stability analysis.}
	To check the stability of the results reported in Table~\ref{tab:main_cifar10_imagenet100}, we repeated RLAT training on CIFAR-10 over 3 different random seeds. The corruption accuracy has the average 84.0\% with the standard deviation of 0.2\% which suggests that the method is quite stable with respect to random seeds.
	
\begin{figure}[t!]
		\begin{center}
			\includegraphics[width=0.36\columnwidth]{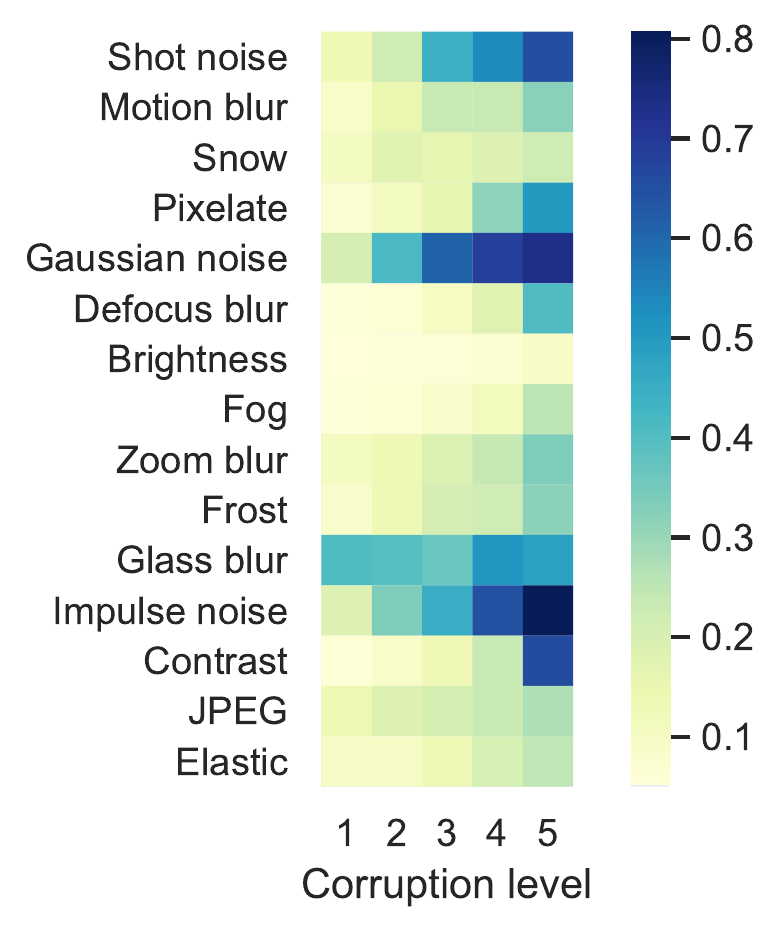}
			\includegraphics[width=0.363\columnwidth]{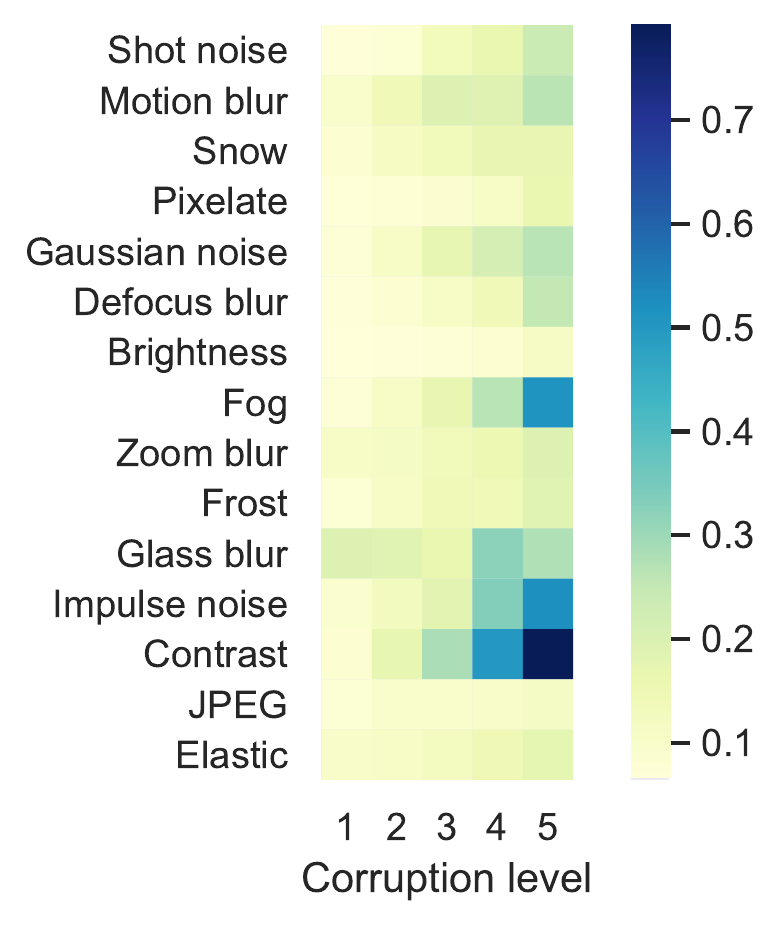}
			\caption{Error rates of a standard and $\l_2$ adversarially trained models kon different common corruptions from CIFAR-10-C.}
			\label{fig:heatmap_err_clean_model_corruptions}
		\end{center}
	\end{figure}

\begin{figure*}[t!]
		\begin{center}
			\includegraphics[width=0.98\columnwidth]{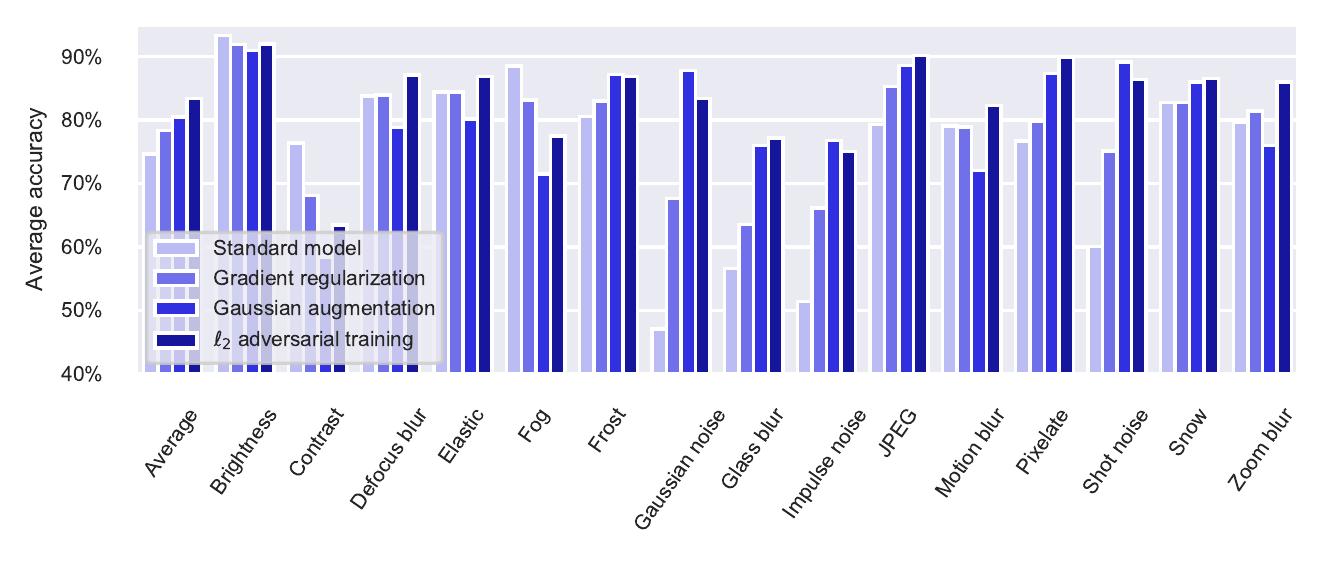}
			\caption{Accuracy for different individual corruptions on CIFAR-10-C. $\l_2$ adversarial training leads to better average performance compared to other baselines.}
			\label{fig:barplot_corruptions}
		\end{center}
	\end{figure*}

	\myparagraph{Performance of different methods across individual corruptions.}
	First, we show the performance of the simple baselines considered in Sec.~\ref{sec:at_vs_gauss} over each of the 15 corruptions of CIFAR-10-C.
	We show in Fig.~\ref{fig:heatmap_err_clean_model_corruptions} the error rates of the standard and the $\l_2$ adversarially trained models and in Fig.~\ref{fig:barplot_corruptions} a breakdown of the accuracy of all simple baselines (i.e., standard and $\l_2$ adversarially trained models, together with the models trained with gradient regularization and Gaussian data augmentation).
	We first note that $\l_2$ adversarial training leads to better average performance compared to other baselines, and improves on each corruption type (blurs, digital, noises, weather) and particularly on JPEG compression, elastic transform, pixelate, and zoom blur. However, compared to the standard model, adversarial training worsens the performance on contrast and fog, and slightly on brightness as has been observed in previous work for $\l_\infty$ adversarial training with a large $\eps=\nicefrac{8}{255}$ \citep{ford2019adversarial}.
	Moreover, Fig.~\ref{fig:heatmap_err_clean_model_corruptions} complements Fig.~\ref{fig:heatmap_distances_corruptions} and helps to motivate why the LPIPS distance can be more suitable than the $\l_2$ distance for the problem of being robust to common corruptions (see the discussion in Sec.~\ref{sec:app_suitability_of_lpips}). 

	We additionally compare these baselines together with 50\% Gaussian  augmentation, AdvProp and RLAT in Table~\ref{tab:clean_gradreg_gauss_adv}. We observe that RLAT leads to better average performance than other baselines (not taking into account 50\% Gaussian augmentation since it is partially trained with a corruption from CIFAR-10-C) and consistently improves upon $\l_2$ adversarial training. AdvProp helps for the snow and elastic transform corruptions, has a better behavior on corruptions on which $\l_2$/RLAT adversarial training and Gaussian augmentation perform poorly (such as brightness, fog, or contrast) but performs suboptimally on noise. Finally, 50\% Gaussian augmentation obtains high accuracy, consistently improving upon 100\% Gaussian augmentation (in particular on blur corruptions) due to its usage of clean samples which mitigates $\sigma$-overfitting. 

	\begin{table*}[t]
		\centering
		\footnotesize
		\begin{tabular}{@{}lccccccc@{}}
			\textbf{Corruption} & \textbf{Clean} & \textbf{GradReg} & \textbf{100\% Gauss} & \textbf{50\% Gauss} & \textbf{AdvProp} & \textbf{$\l_2$ AT} & \textbf{RLAT} \\
			\midrule
			Shot noise & 60.1\% & 75.1\% & 89.2\% &   \textbf{91.2\%} & 82.0\%       & 86.4\%& {88.8\%}\\
			Motion blur & 79.1\% & 78.9\% &          72.0\% &  82.2\%         & 82.2\%&      \textbf{82.3\%} & \textbf{ 82.3\%}  \\
			Snow & 82.8\% & 82.8\%     &    86.0\% &                85.9\%  & \textbf{86.9}\% &86.6\% & 86.2\% \\
			Pixelate & 76.7\% & 79.8\% &    87.4\% &    87.7\%  &  87.7\%       &  89.9\%   &\textbf{90.2\%} \\
			Gaussian noise & 47.1\% & 67.6\% &   \color{gray} 87.8\% &  \color{gray} 90.8\%       & 77.0\%      &   83.4\% & \textbf{86.0}\%  \\
			Defocus blur & 83.8\% & 83.9\% &          78.8\% &        86.5\%  & {87.1\%}    &   {87.1\%} & \textbf{87.2\%} \\
			Brightness & \textbf{93.3\%} & 91.9\% & 91.0\% &   91.2\%   & 92.8\%       &92.0\% & 91.5\%\\
			Fog & \textbf{88.5\%} & 83.1\% & 71.5\% &       78.7\%    &88.2\%    &  77.5\% &  76.7\% \\
			Zoom blur & 79.7\% & 81.4\% &          76.0\% &       85.2\%    & 85.8\%     &  {86.0\%}& \textbf{86.1\%} \\
			Frost & 80.6\% & 83.0\% &         \textbf{87.2\%} &         \textbf{87.2\%}  & 86.4\%    & 86.9\% & 87.0\%  \\
			Glass blur & 56.6\% & 63.5\% &   76.0\% &   79.3\%      & 74.4\%   & {77.2\%}&  \textbf{80.4\%}  \\
			Impulse noise & 51.4\% & 66.1\%  &       76.8\% &      \textbf{87.3\%}   & 71.8\%       &  75.1\%& {79.6\%}   \\
			Contrast & \textbf{76.4\%} & {68.1\%}    &     58.4\% &         66.3\%  & 68.6\%   &  63.5\%&62.6\%  \\
			JPEG  & 79.3\% & 85.3\%   &   88.6\% &           89.8\%  & 89.8\%   &  {90.2\%}& \textbf{90.5\%} \\
			Elastic  & 84.4\% & 84.4\%   &   80.2\% &              86.3\% & \textbf{87.9\%} &  {86.9\%}& {87.2\%}   \\
			\midrule
			Average &   74.6\% &    78.3\%  &   \color{gray} 80.5\% &  \color{gray} 85.0\%  & 82.9\%      &  {83.4\%}& \textbf{84.1\%} \\
		\end{tabular}
		\caption{Accuracy of clean training, gradient regularization, 100\% and 50\% Gaussian data augmentation, AdvProp, $\ell_2$ adversarial training, and RLAT on CIFAR-10-C using ResNet-18. Gray-colored numbers correspond to methods that were at least partially trained with the corruptions from CIFAR-10-C.}
		\label{tab:clean_gradreg_gauss_adv}
	\end{table*}

	\begin{table*}[t!]
		\centering
		\setlength{\tabcolsep}{4.3pt}
		\footnotesize
		\begin{tabular}{@{}l ccccccccc@{}}
			& \multicolumn{9}{c}{\textbf{Radius $\eps$ used for $\l_2$ adversarial training}} \\
			\cmidrule(lr){2-10}
			\textbf{Corruption} & $\varepsilon=0$   &$\varepsilon=0.01$   & $\varepsilon=0.05$  & $\varepsilon=0.08$   & $\varepsilon=0.1$  & $\varepsilon=0.15$  & $\varepsilon=0.2$  & $\varepsilon=0.5$  & $\varepsilon=1$  \\
			
			\midrule
			Shot noise        & 60.1\% & 70.6\% & 81.6\% & 84.9\%  & 86.4\%  &	\textbf{87.0\%} & 86.9\%  & 85.1\%  & 79.6\%  \\
			Motion blur       & 79.1\% & 79.4\% & 81.4\% & 81.8\%  &	\textbf{82.3\%} & 82.1\%  & 82.0\%  & 80.0\%  & 75.3\%  \\
			Snow              & 84.7\% & 82.8\% & 86.6\% &\textbf{86.9\%} & 86.6\%  & 86.0\%  & 85.5\%  & 81.3\%  & 75.0\% \\
			Pixelate          & 76.7\% & 82.0\% & 88.8\% & 89.7\% & 89.9\% &	\textbf{90.2\%}& 89.7\% & 86.5\% & 81.2\% \\
			Gaussian noise    & 47.1\% & 59.6\% & 76.0\%  & 81.3\% & 83.4\% & 84.6\% &	\textbf{85.1\%}& 83.8\% & 78.4\%  \\
			Defocus blur      & 83.8\% & 84.5\% & 86.2\%  & 86.7\% &	\textbf{87.1\%}& 86.8\% & 86.5\% & 83.7\% & 78.6\% \\
			Brightness        & \textbf{93.3\%} & 93.2\% & 93.0\% & 92.5\% & 92.0\% & 91.3\% & 90.2\% & 85.4\% & 79.0\%  \\
			Fog               & \textbf{88.5\%} & 86.7\% &80.8\%& 78.5\% & 77.5\% & 74.4\% & 71.5\% & 61.5\% & 54.1\%  \\
			Zoom blur         & 79.7\% & 81.9\% & 84.7\% & 85.6\% &	\textbf{86.0\%}& 85.7\% & 85.4\% & 82.7\% & 77.8\%  \\
			Frost             & 80.6\% & 84.4\% & \textbf{87.5\%}&	\textbf{87.5\%}& 86.9\% & 86.4\% & 85.7\% & 79.7\% & 71.5\%  \\
			Glass blur        & 56.6\% & 62.7\% & 72.9\% & 76.1\% & 77.2\% & 80.7\% &	\textbf{81.6\%}& 81.5\% & 76.7\%  \\
			Impulse noise     & 51.5\% & 58.7\% & 66.4\% & 73.2\% & 75.1\% & 76.5\% & 78.1\% &	\textbf{79.7\%}& 75.0\%  \\
			Contrast          & \textbf{76.4\%} & 72.3\%& 65.9\% & 63.7\% & 63.5\% & 61.1\% & 59.3\% & 50.5\% & 44.0\%  \\
			JPEG              & 79.3\% & 86.0\% & 89.9\% &	\textbf{90.4\%}& 90.2\% & 90.3\% & 89.9\% & 86.7\% & 81.5\%       \\
			Elastic           & 84.4\% & 85.7\% & \textbf{87.2\%}&	\textbf{87.2\%} & 86.9\% & 86.6\% & 85.9\% & 82.4\% & 77.1\%  \\
			\midrule
			Average           & 74.6\% &   78.2\%& 81.9\% & 83.1\% &	\textbf{83.4\%}& 83.3\% & 82.9\% & 79.4\% & 73.7\%  \\
		\end{tabular}
		\caption{Accuracy of $\ell_2$ adversarial training for different $\varepsilon$ on CIFAR-10-C. Similarly to Fig.~\ref{fig:linf_eps_c10c_acc} which was done for $\l_\infty$ norm, we observe here that the widely used $\l_2$ radius $\eps=0.5$ leads to suboptimal corruption accuracy.}
		\label{tab:l2_at_many_eps_table}
	\end{table*}

	\myparagraph{Detailed performance of $\l_2$ adversarial training trained with different $\eps$.}
	We study further the influence of the radius $\eps$ on the performance of $\l_2$ adversarial training on CIFAR-10-C.
	We present in Table~\ref{tab:l2_at_many_eps_table} the accuracy of $\l_2$ adversarially trained model, trained with different values of $\eps$. 
	Interestingly, Table~\ref{tab:l2_at_many_eps_table} suggests that the same degradation for fog and contrast occurs even for the \textit{smallest} $\eps$ used for training. 
	We also observe that different corruptions require different $\eps$ for optimal performance.
	In particular, the optimal $\eps$ for the frost corruptions is $0.05$ while for impulse noise the optimal $\eps$ is an order of magnitude larger, i.e. $0.5$. Despite being widely used in the literature on $\l_2$ robustness, $\eps=0.5$ is suboptimal for many other corruptions, and the optimal average performance over \textit{all} corruptions for a single model is obtained at $\eps=0.1$.

	\myparagraph{Detailed model comparison for different corruption severity.}
	In Tables~\ref{tab:app_main_cifar10_severity} and~\ref{tab:app_main_imagenet100_severity} we report the accuracy of selected models for each corruption level separately. 
	We first note that for all models, as one could expect, the accuracy is gradually decreasing with the corruption severity. 
	When comparing all the baselines, RLAT shows the best accuracy on the majority of corruption levels (three on Imagenet-100 and four on CIFAR-10). It is also worth mentioning that AdvProp on CIFAR-10 works better for smaller corruption levels which can be explained by its higher standard accuracy. 
	On ImageNet-100-C, we can observe that the Fast PAT model has the best accuracy at the highest severity level despite having clearly suboptimal standard accuracy (71.5\% compared to 86.6\% of the standard model) and average corruption accuracy (45.2\% compared to 47.5\% of the standard model). We also observe that for the severity level 4, $\l_2$ adversarial training is slightly better than RLAT (37.0\% vs 36.8\%).
	Moreover, we notice that when RLAT is combined with different data augmentation schemes, the improvement is often achieved at multiple severity levels simultaneously. For example, when RLAT is combined with AugMix, the accuracy is improved on all severity levels.

	\begin{table*}[t!]
		\centering
		\footnotesize
		\begin{tabular}{@{}lrrrrrrr@{}}
			&           & \multicolumn{6}{c}{\textbf{Accuracy at different severity levels}} \\
			\cmidrule(lr){3-8}
			\textbf{Method} &     Standard &     1 &     2 &     3 &     4 &     5 &   Average \\
			\midrule
			Standard training   & 95.1\% & 87.6\% & 82.3\% & 76.2\% & 69.2\% & 57.9\% & 74.6\% \\
			Gradient regularization  & 93.4\% & 87.9\% & 84.9\% & 81.0\% & 74.8\% & 65.5\% & 78.8\% \\
			100\% Gaussian augmentation & 92.5\% & \color{gray} 89.2\% & \color{gray} 86.3\% & \color{gray} 82.3\% & \color{gray} 76.4\% & \color{gray} 68.1\% & \color{gray} 80.5\% \\ \color{black}
			50\%  Gaussian augmentation & 93.2\% & \color{gray} 91.0\% & \color{gray} 89.1\% & \color{gray} 86.8\% & \color{gray} 82.6\% & \color{gray} 75.8\% & \color{gray} 85.0\% \\ \color{black}
			Fast PAT & 93.4\% & 89.5\% & 87.2\% & 83.7\% & 79.0\% & 72.6\% & 82.4\% \\
			$\l_\infty$ AT & 93.3\% & 90.8\% & 88.3\% & 84.5\% & 78.5\% &  71.2\% & 82.7\% \\
			AdvProp    & 94.7\% & \textbf{91.5}\% & 88.9\% & 85.2\% & 79.2\% & 69.5\% & 82.9\% \\
			$\l_2$ AT  & 93.6\% & 91.1\% & 88.8\% & 85.5\% & 79.8\% & 71.8\% & 83.4\% \\
			RLAT  & 93.1\% & 91.1\% & \textbf{89.0\%} & \textbf{85.9\%} & \textbf{80.9\%} & \textbf{73.5\%} & \textbf{84.1\%} \\
			\midrule
			DeepAugment & 94.1\% & 91.0\% & 89.0\% & 86.6\% & 82.7\% & 77.3\% & 85.3\% \\
			DeepAugment + RLAT & 93.6\% & \textbf{91.7\%} & \textbf{90.6\%} & \textbf{89.1\%} & \textbf{86.1\%} & \textbf{81.6\%} & \textbf{87.8\%} \\
			\midrule
			AugMix  & 95.0\% & 92.2\% & 90.5\% & 88.5\% & 84.7\% & 78.8\% & 86.9\% \\
			AugMix + RLAT & 94.8\% & \textbf{93.1\%} & \textbf{91.8\%} & \textbf{90.3\%} & \textbf{87.0\%} & \textbf{80.6\%} & \textbf{88.5\%} \\
			\midrule
			AugMix + JSD & 95.0\% & 92.9\% & 91.5\% & 89.9\% & 86.8\% & 82.1\% & 88.6\% \\
			AugMix + JSD + RLAT & 94.8\% & \textbf{93.3\%} & \textbf{92.3\%} & \textbf{90.9\%} & \textbf{88.3\%} & \textbf{83.3\%} & \textbf{89.6\%} \\
		\end{tabular}
		\caption{Accuracy of different methods on CIFAR-10-C. Gray-colored numbers correspond to methods that were at least partially trained with the corruptions from CIFAR-10-C.}
		\label{tab:app_main_cifar10_severity}
	\end{table*}
	
	\begin{table*}[t!]
		\centering
		\footnotesize
		\begin{tabular}{@{}lrrrrrrr@{}}
			&           & \multicolumn{6}{c}{\textbf{Accuracy at different severity levels}} \\
			\cmidrule(lr){3-8}
			\textbf{Method} &     Standard &     1 &     2 &     3 &     4 &     5 &   Average \\
			\midrule
			Standard training & 86.6\% & 70.9\% & 58.7\% & 47.3\% & 35.2\% & 25.4\% & 47.5\% \\ 
			100\% Gaussian augmentation & 86.4\% & \color{gray} 70.1\% & \color{gray} 57.2\% & \color{gray} 46.2\% & \color{gray} 34.9\% & \color{gray} 25.3\% & \color{gray} 46.7\% \\ \color{black}
			50\%  Gaussian augmentation & 83.8\% & \color{gray} 73.8\% & \color{gray} 65.0\% & \color{gray} 56.9\% & \color{gray} 45.7\% & \color{gray} 34.8\% & \color{gray} 55.2\% \\ \color{black}
			Fast PAT & 71.5\% & 61.7\% & 52.7\% & 45.7\% & 36.6\% & \textbf{29.1}\% & 45.2\%\\
			$\l_\infty$ AT   & 86.5\% & 70.6\% & 57.9\% & 46.5\% & 36.2\% & 27.4\% & 47.7\% \\
			$\l_2$ AT       &  86.3\%  & 70.1\% & 58.3\% & 47.8\% & \textbf{37.0\%} & 27.9\% & 48.4\% \\
			RLAT & 86.5\% & \textbf{71.6\%} & \textbf{59.6\%} & \textbf{48.8\%} & 36.8\% & 27.1\% & \textbf{48.8\%} \\
			\midrule
			AugMix & 86.7\% & 73.9\% & 63.3\% & 53.9\% & 41.1\% & 29.5\% & 52.3\% \\ 
			AugMix + RLAT & 86.8\% & \textbf{75.4\%} & \textbf{65.1\%} & \textbf{56.2\%} & \textbf{44.7\%} & \textbf{32.8\%} & \textbf{54.8\%} \\  
			\midrule
			Stylized ImageNet & 86.6\% & 73.0\% & 63.5\% & 55.1\% & 43.5\% & \textbf{33.2\%} & 53.7\% \\  
			Stylized ImageNet + RLAT & 86.5\% & \textbf{74.1\%} & \textbf{64.4\%} & \textbf{56.0\%} & \textbf{44.1\%} & 33.0\% & \textbf{54.3\%} \\   
			\midrule
			ANT\textsuperscript{3x3} & 85.9\% & 74.2\% & \textbf{66.5\%} & \textbf{58.9\%} & 49.8\% & 39.3\% & 57.7\% \\ 
			ANT\textsuperscript{3x3} + RLAT & 85.3\% & \textbf{74.5\%} & 66.4\% & 58.8\% & \textbf{50.5\%} & \textbf{41.2\%} & \textbf{58.3\%} \\ 
		\end{tabular}
		\caption{Accuracy of different methods on ImageNet-100-C. Gray-colored numbers correspond to methods that were at least partially trained with the corruptions from ImageNet-100-C.}
		\label{tab:app_main_imagenet100_severity}
	\end{table*}

	\myparagraph{Results of AugMix combined with $\l_2$ and $\l_\infty$ adversarial training.}
	We report these results on CIFAR-10 in Table~\ref{tab:augmix_at_extra}. We can observe that RLAT still outperforms $\l_2$ and $\l_\infty$ adversarial training when they are combined with AugMix, both with and without the JSD consistency term from \cite{hendrycks2019augmix}.
	Moreover, our RLAT model achieves 89.4\% accuracy with the JSD term outperforming the best model from the AugMix paper \cite{hendrycks2019augmix} despite being much smaller: we use ResNet-18 while the best model of \cite{hendrycks2019augmix} uses ResNeXt-29.
	\begin{table}[t]
		\centering
		\small
		\begin{tabular}{@{}lc@{}}
			\textbf{Method} &  \textbf{Accuracy} \\
			\midrule
			AugMix + $\l_\infty$ adversarial training & 87.8\% \\  
			AugMix + $\l_2$ adversarial training      & 88.3\% \\  
			AugMix + RLAT           & \textbf{88.5\%} \\
			\midrule
			AugMix + JSD + $\l_\infty$ adversarial training & 89.0\% \\
			AugMix + JSD + $\l_2$ adversarial training      & 89.0\% \\
			AugMix + JSD + RLAT                             & \textbf{89.6\%}
		\end{tabular}
		\caption{CIFAR-10-C accuracy of AugMix without and with the JSD term combined with methods based on adversarial training ($\l_\infty$, $\l_2$, RLAT).}
		\label{tab:augmix_at_extra}
	\end{table}

	\myparagraph{Model selection based on validation corruptions.}
	Many previous works use the same model selection scheme as we used, i.e. selecting the optimal hyperparameters on the test corruptions \cite{hendrycks2019augmix, xie2020adversarial, rusak2020simple}. However, a proper validation-test split would lead to a more rigorous evaluation which we report in Table~\ref{tab:app_model_selection_c10c_extra}. We do a grid search based on the accuracy on the four validation corruptions from CIFAR-10-C. We see that for all adversarially trained models, \textit{the optimal $\eps$ is the same} as we reported in Table~\ref{tab:main_cifar10_imagenet100} except for the $\l_\infty$ model on ImageNet-100 for which, however, the final test accuracy is unchanged. Overall, this is perhaps not too surprising as the validation corruptions are from the same category of corruptions and we optimize over a small one-dimensional grid so we are unlikely to overfit to the test corruptions in this way.
	\begin{table}[t]
		\centering
		\small
		\begin{tabular}{@{}lccc@{}}
			\textbf{Method} &  \textbf{Test accuracy}              & \textbf{Validation accuracy} & \textbf{Same $\pmb{\eps}$?} \\
			\midrule
			$\l_\infty$ AT & 82.7\% / 47.7\%                   & 86.6\% / 53.2\%                   & \textbf{Yes} / No \\  
			$\l_2$ AT      & 83.4\% / 48.4\%                   & 86.9\% / 54.0\%                   & \textbf{Yes} / \textbf{Yes}   \\
			RLAT           & \textbf{84.1\%} / \textbf{48.8\%} & \textbf{87.1\%} / \textbf{54.7\%} & \textbf{Yes} / \textbf{Yes}  \\
		\end{tabular}
		\caption{Accuracy on the 15 test and 4 extra validation corruptions from \textit{CIFAR-10-C} / \textit{ImageNet-100-C} for the best models out of a grid of $\eps$ values for each method.}
		\label{tab:app_model_selection_c10c_extra}
	\end{table}

	\myparagraph{Results on larger architectures.}
	Since we performed all experiments for Table~\ref{tab:app_model_selection_c10c_extra} on ResNet-18, we also test here another architecture to make sure that our findings generalize to other models.
	For this, we take the WRN-28-10 architecture on CIFAR-10 and similarly to other experiments, perform a grid search over $\eps$ for adversarial training methods ($\l_\infty$, $\l_2$, RLAT). We report the results on the best $\eps$ for each method in Table~\ref{tab:app_wrn_exps} and observe that RLAT also outperforms other adversarial training methods for this architecture.
	\begin{table}[t!]
		\centering
		\small
		\begin{tabular}{@{}lc@{}}
			\textbf{Method} &  \textbf{Accuracy} \\
			\midrule
			Standard training & 75.6\% \\
			$\l_\infty$ adversarial training & 84.8\% \\  
			$\l_2$ adversarial training      & 85.5\% \\  
			RLAT           & \textbf{85.9\%}
		\end{tabular}
		\caption{CIFAR-10-C accuracy of adversarial training methods ($\l_\infty$, $\l_2$, RLAT) using WRN-28-10 architecture.}
		\label{tab:app_wrn_exps}
	\end{table}

	\myparagraph{Calibration before and after temperature rescaling.}
	In Table~\ref{tab:app_cifar10_imagenet100_calibration} we report ECE before and after calibration via temperature rescaling \citep{guo2017calibration} using in-distribution data for the models reported in Table~\ref{tab:main_cifar10_imagenet100}. 
	We observe that calibration is substantially improved after temperature rescaling, however it does not affect the ranking between different methods. In particular, all adversarial training methods significantly improve the calibration error, and RLAT leads to the best calibration reaching $1.3\%$ ECE on CIFAR-10-C when combined with AugMix.
	Moreover, we note that all the CIFAR-10 models are \textit{overconfident} since their optimal calibration temperature is greater than $1$ (mostly in the range from $1.2$ to $1.5$).
	On ImageNet-100, the picture is similar and RLAT also improves calibration. For example, ANT\textsuperscript{3x3} leads to 4.5\% ECE after calibration while ANT\textsuperscript{3x3} combined with RLAT achieves 2.8\% ECE.
	\begin{table}[t!]
		\centering
		\small
		\setlength{\tabcolsep}{4.5pt}
		\begin{tabular}{@{}lcc}
			& \textbf{ECE before}  & \textbf{ECE after}    \\
			\textbf{Training} & \textbf{calibration} & \textbf{calibration}  \\
			\midrule
			\\
			& \multicolumn{2}{c}{\textbf{CIFAR-10-C}}  \\
			\cmidrule(lr){2-3}
			Standard & 16.6\% & 11.3\%                                 \\
			100\% Gaussian & \color{gray} 13.2\% & \color{gray} 7.8\%  \\ 
			50\%  Gaussian & \color{gray} 9.1\% & \color{gray} 4.6\%   \\ 
			Fast PAT & 12.0\% & 6.6\%                                  \\
			AdvProp  & 10.1\% & 6.4\%                                  \\
			$\l_\infty$ adversarial  & 10.8\% & 6.5\%                  \\
			$\l_2$ adversarial & 10.5\% & 5.8\%                        \\
			RLAT & \textbf{9.9\%} & \textbf{5.1\%}                     \\
			\midrule
			DeepAugment         & 8.7\% & 4.4\%                        \\  
			DeepAugment + RLAT  & \textbf{6.1\%} & \textbf{2.3\%}      \\  
			\midrule
			AugMix & 6.9\% & 3.2\%                                     \\ 
			AugMix + RLAT & \textbf{4.5\%} & \textbf{1.3\%}            \\ 
			\midrule
			AugMix + JSD & 6.5\% & 4.2\%                               \\
			AugMix + JSD + RLAT & \textbf{5.4\%} & \textbf{3.3\%}      \\  
			\\
			& \multicolumn{2}{c}{\textbf{ImageNet-100-C}} \\
			\cmidrule(lr){2-3}
			Standard        & 10.0\% & 6.4\% \\
			100\% Gaussian  & \color{gray} 11.7\% & \color{gray} 8.9\% \\ 
			50\%  Gaussian  & \color{gray} 6.1\% & \color{gray} 4.5\% \\ 
			Fast PAT        & 8.0\% & 12.7\% \\
			$\l_\infty$ adversarial                  & 12.4\% & 10.9\% \\
			$\l_2$ adversarial                       & 9.4\% & 6.5\% \\
			RLAT                                     & \textbf{9.1\%} & \textbf{5.4\%} \\
			\midrule
			AugMix            & 7.5\% & 5.8\% \\ 
			AugMix + RLAT     & \textbf{4.7\%} & \textbf{5.3\%} \\ 
			\midrule
			AugMix + JSD      & 1.9\% & 4.0\% \\
			AugMix + JSD + RLAT   & \textbf{1.8\%} & \textbf{2.1\%} \\  
			\midrule
			SIN          & 6.7\% & 5.8\% \\  
			SIN + RLAT   & \textbf{6.0\%} & \textbf{5.1\%} \\  
			\midrule 
			ANT\textsuperscript{3x3}           & 5.1\% & 4.5\% \\ 
			ANT\textsuperscript{3x3} + RLAT    & \textbf{4.4\%} & \textbf{2.8\%} \\
		\end{tabular}
		\caption{Calibration of ResNet-18 models trained on CIFAR-10 and ImageNet-100 before and after calibration via temperature rescaling. Gray-colored numbers correspond to methods partially trained with the corruptions from CIFAR-10-C and ImageNet-100-C.}
		\label{tab:app_cifar10_imagenet100_calibration}
	\end{table}

	\myparagraph{Comparison of in-distribution vs. out-distribution calibration.}
	Throughout the paper, we focused only on calibration and accuracy on \textit{common image corruptions}. Regarding calibration on in-distribution images, it is known that $\ell_p$ adversarial training can degrade the calibration quality (e.g., see \citet{croce2020robustbench}). However, with the small $\eps$ that we use for adversarial training, this degradation is minimal. To illustrate this, we report below the calibration results on clean and corrupted CIFAR-10 data in Table~\ref{tab:app_calibration_in_vs_out}.
	We can see that, in line with the literature, $\ell_p$ adversarial training degrades the calibration on clean data but only slightly (from 2.9\% to 4.0\% for RLAT). At the same time, all adversarial training methods significantly improve the calibration on corrupted data (from 16.6\% to 9.9\% for RLAT).
	\begin{table}[t]
		\centering
		\small
		\begin{tabular}{@{}lcc@{}}
			& \textbf{ECE on} & \textbf{ECE on} \\
			\textbf{Method} & \textbf{CIFAR-10} & \textbf{CIFAR-10-C} \\
			\midrule
			Standard training                & \textbf{2.9\%} & 16.6\% \\
			$\l_\infty$ adversarial training & 3.9\%          & 10.8\% \\  
			$\l_2$ adversarial training      & 3.7\%          & 10.5\% \\  
			RLAT                             & 4.0\%          & \textbf{9.9\%} \\
		\end{tabular}
		\caption{A comparison of the expected calibration error (ECE) on in-distribution (CIFAR-10) vs. out-distribution (CIFAR-10-C) data.}
		\label{tab:app_calibration_in_vs_out}
	\end{table}

\end{document}